\documentclass{article}

\PassOptionsToPackage{numbers,square,comma}{natbib}
 \usepackage[preprint]{neurips_2026}

\usepackage[utf8]{inputenc} % allow utf-8 input
\usepackage[T1]{fontenc}    % use 8-bit T1 fonts
\usepackage{url}            % simple URL typesetting
\usepackage{booktabs}       % professional-quality tables
\usepackage{amsfonts}       % blackboard math symbols
\usepackage{nicefrac}       % compact symbols for 1/2, etc.
\usepackage{microtype}      % microtypography
\usepackage{xcolor}         % colors
\definecolor{citeblue}{HTML}{1A0DAB}  % 深蓝色，可自行调整
\usepackage[
    colorlinks=true,
    linkcolor=citeblue,
    citecolor=citeblue,
    urlcolor=citeblue
]{hyperref}       % hyperlinks
\usepackage{graphicx}
\usepackage{amsmath}
\usepackage{multirow}
\usepackage{caption}
\usepackage[table]{xcolor}
\usepackage{subcaption}
\newcommand{\same}{\ensuremath{\dagger}}
\usepackage{float}
\usepackage{placeins}
\usepackage{wrapfig}
\usepackage{caption}
\usepackage{tabularx}

\title{ReSCUE: Re-translation with Sentence Commitment for Unsegmented Long-Form Simultaneous Sign Language Translation}

\author{
\textbf{Sihan Ren}\textsuperscript{*,1,3,$\dagger$} \quad
\textbf{Gaozheng Li}\textsuperscript{*,1,3,$\dagger$} \quad
\textbf{Yuanshang Quan}\textsuperscript{1,3,$\dagger$} \quad
\textbf{Yiming Qin}\textsuperscript{1,3,$\dagger$} \\
\textbf{Fuyi Yang}\textsuperscript{1,3,$\dagger$} \quad
\textbf{Chang Liu}\textsuperscript{1,3,$\dagger$} \quad
\textbf{Lan Xu}\textsuperscript{1} \quad
\textbf{Minye Wu}\textsuperscript{2} \\[1ex]
\textsuperscript{1}ShanghaiTech University
\qquad
\textsuperscript{2}University of Derby
\qquad
\textsuperscript{3}DGene
}

\begin{document}

\maketitle

\begingroup
\renewcommand{\thefootnote}{}
\footnotetext{\textsuperscript{*}Equal contribution. \quad
\textsuperscript{$\dagger$}Work done during an internship at DGene.}
\addtocounter{footnote}{-1}
\endgroup

\begin{abstract}
Simultaneous Sign Language Translation (SLT) is critical for real-time communication, yet existing methods remain largely confined to sentence-level, offline settings that assume pre-segmented inputs. These assumptions hinder deployment in realistic scenarios involving continuous, unsegmented video streams. 
% %
We present ReSCUE, a unified framework for simultaneous SLT on unsegmented long-form sign language videos that aligns training and inference with realistic streaming conditions. ReSCUE combines inference-aware training to handle partial inputs, non-signing pauses, and multi-sentence contexts, stabilized re-translation to enable low-latency yet revisable predictions with reduced output flicker, and a sentence commitment mechanism for online segmentation and memory management. 
% %
Experiments on standard sentence-level benchmarks show that ReSCUE achieves lower latency and the best translation quality under low-latency settings. On long-form unsegmented datasets, ReSCUE approaches the translation quality of oracle offline systems that use ground-truth sentence boundaries, while operating at substantially lower latency, demonstrating its practicality for real-world streaming scenarios.
\end{abstract}

\section{Introduction}

Sign Language Translation (SLT) is a critical technology for enabling effective communication between Deaf and hearing communities and for supporting the social inclusion of deaf individuals in everyday life~\citep{camgoz2018neural, zhou2021improving, li2025uni}.
However, many real-world scenarios, such as live presentations, meetings, classrooms, and public events, remain beyond the reach of existing SLT systems, as they are not designed to support continuous, long-duration interaction.
To make SLT practical in these cases, translation needs to be performed simultaneously on unsegmented long-form video streams.

Recent years have witnessed rapid progress in SLT, driven by stronger architectures, large-scale pretraining, better utilization of auxiliary information like context, and the integration of large language models, leading to improving translation quality in offline settings~\cite{li2025uni,wong2024sign2gpt,zhao2024conditional,jang2025lost}.
Yet, most existing SLT methods remain focused on sentence-level, offline translation, assuming access to the complete video before inference begins.
While a few studies have explored simultaneous SLT~\cite{simulslt,sun-etal-2024-adaptive,zuo2024towards}, they are still restricted to sentence-level settings and rely on pre-segmented inputs, leaving long-form, unsegmented scenarios largely unexplored.
In contrast, simultaneous translation has been extensively studied and deployed in other modalities, such as text~\cite{ma2019stacl,mamonotonic,zheng-etal-2020-simultaneous} and speech~\cite{arivazhagan2020retranslationversusstreamingsimultaneous,zeng2021realtrans,zhang-etal-2024-streamspeech}, where mature methods successfully balance translation quality, latency, and output stability. Insights have not yet been fully explored in SLT. 
Besides, despite recent advances on sentence-level SLT datasets~\cite{zhou2021improving,camgoz2018neural,how2sign2021,uthus2023youtube,bobsl2021} with larger scale, broader topical coverage, richer modalities, their pre-segmented single-sentence format limits training and evaluation on long-form, unsegmented videos.

{
\setlength{\textfloatsep}{6pt}
\setlength{\intextsep}{5pt}
\begin{figure}[H]
  \centering
\includegraphics[width=\textwidth,height=0.3\textheight,keepaspectratio]{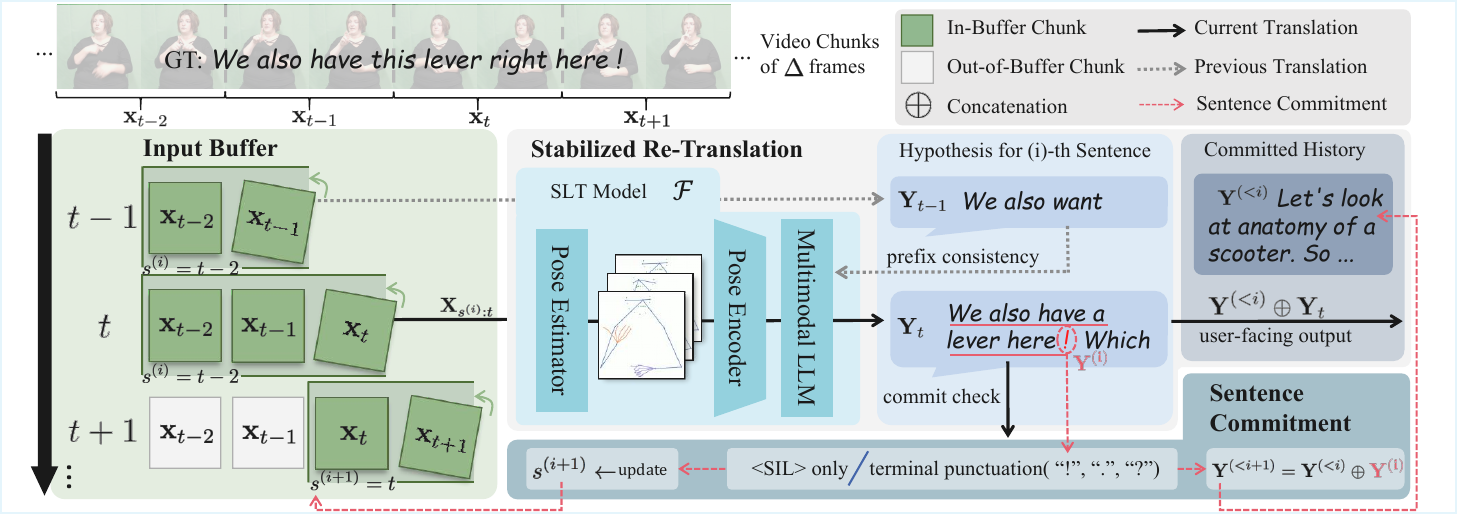}
  \caption{
Overview of the proposed simultaneous SLT pipeline.
The unsegmented video stream is processed in fixed-length chunks and accumulated in a dynamic input buffer.
At each step, the SLT module performs re-translation over the buffered video segment
to produce a revisable hypothesis for the current sentence, while encouraging prefix consistency with the previous hypothesis.
Finalized translations of previous sentences are stored in the committed history and concatenated with the current hypothesis to form the user-facing output.
A sentence commitment mechanism detects sentence completion, extracts the finalized translation, appends it to the committed history, and advances the input buffer accordingly.
}
  \label{fig:overview}
\end{figure}
}

In this work, we take a first step toward practical simultaneous SLT on unsegmented long-form videos. 
We train SLT models under realistic streaming conditions by exposing them to partial inputs, non-signing pauses, and consecutive sentences, enabling robust translation in real-world scenarios. 
We further adapt re-translation~\cite{arivazhagan2020retranslationversusstreamingsimultaneous} to SLT to achieve high-quality, low-latency simultaneous translation, and an inference-time sentence commitment mechanism that automatically segments continuous video streams and manages memory during inference.

% As illustrated in~\ref{fig:overview}, ReSCUE comprises three components: \emph{Inference-Aware Training} constructs prefix-to-prefix, silence-augmented, and multi-sentence training instances to align training with streaming inference.
% \emph{Stabilized Re-Translation} adapts re-translation to SLT using biased search and Mask-$k$ to generate stable hypotheses in a chunk-by-chunk manner.
% \emph{Sentence Commitment} detects sentence completion during inference, commits finalized translations, and advances the input buffer to segment continuous video streams.

As illustrated in Fig.~\ref{fig:overview}, ReSCUE integrates three tightly coupled components to enable robust simultaneous translation on continuous video streams.
To expose the model to realistic streaming conditions, \emph{Inference-Aware Training} constructs prefix-to-prefix, silence-augmented, and multi-sentence training instances, allowing the model to handle partial inputs, pauses, and consecutive sentences during inference.
To produce stable and low-latency outputs as visual context evolves over time, \emph{Stabilized Re-Translation} adapts re-translation to SLT using biased search and Mask-$k$, thereby suppressing unstable revisions in chunk-by-chunk translation.
To operate directly on unsegmented long-form videos where sentence boundaries are not available in real-world settings, \emph{Sentence Commitment} incrementally detects sentence completion from streaming hypotheses, commits finalized translations once sufficient evidence is observed, and advances the input buffer to support continuous translation without requiring pre-defined segmentation.

We evaluate our method on both sentence-level and long-form settings. On CSL-Daily~\cite{zhou2021improving} and Phoenix-2014-T~\cite{camgoz2018neural}, our approach establishes a new quality–latency Pareto frontier among existing simultaneous SLT methods.
To assess performance in realistic scenarios, we introduce two long-form unsegmented datasets, How2Sign-Long and CSL-Story, where each video contains on average 14 and 63 sign language sentences, respectively.
Experiments on these datasets validate the effectiveness of our method on long-form sign language videos, and we further demonstrate stable real-time translation through a demo video and a user study.

We envision this work as a step toward moving Simultaneous SLT beyond sentence-level settings and toward realistic long-form streaming scenarios.
ReSCUE provides a practical path for deploying SLT systems in real-world interactive environments.
Beyond technical contributions, robust simultaneous SLT is pivotal for dismantling communication barriers, empowering the Deaf community to fully integrate into society, and fostering a more inclusive world.

\section{Related Work}
\subsection{Offline Sign Language Translation} 
Sign Language Translation (SLT) aims to translate sign videos into spoken language text. 
Early approaches typically adopt a cascaded pipeline, where Sign Language Recognition (SLR) first produces gloss sequences, which are then translated into text by a Machine Translation (MT) model \cite{camgoz2018neural,  zhou2021improving, chen2022two,liu2025scope}. 

Recent works adopt direct video-to-text models for SLT \cite{li2025uni, zhou2023gloss, chen2025c, gong2024llms}, which reduce system latency by eliminating intermediate recognition stages and, with large-scale pretraining and multimodal language models, achieve strong offline translation performance. However, these methods typically assume pre-segmented inputs and require full-video access before inference, limiting their use in real-time settings.

\subsection{Simultaneous Translation}
Simultaneous translation generates outputs concurrently with a growing input stream.
Mainstream approaches fall into two categories: \textit{incremental decoding} (e.g., Wait-$k$) \cite{dalvi2018incrementaldecodingtrainingmethods, ma2020simulmt, elbayad2020efficient}, which produces monotonic outputs with fixed latency; and \textit{re-translation} \cite{niehues2018low, arivazhagan2020retranslationversusstreamingsimultaneous, ArivazhaganRetrans, sen2023self}, which continuously refreshes the generated hypothesis as new input arrives, enabling the correction of temporary errors caused by insufficient context.
Some prior work also explores commitment mechanisms to control the input context and determine which outputs should no longer be revised.
These methods either rely on auxiliary intermediate signals or extra modules~\cite{Niehues2016DynamicTF, yao2020dynamic, ArivazhaganRetrans}, or use a special boundary token as a commitment signal~\cite{iranzo-sanchez-etal-2024-segmentation}.

While recent works have extended these paradigms to Simultaneous SLT \cite{simulslt, sun-etal-2024-adaptive, zuo2024towards}, they mostly cannot revise emitted tokens and remain limited to pre-segmented, sentence-level inputs, failing to address the challenges of unsegmented, long-form streams (e.g., boundary detection) in real-world deployment.
% they remain limited in two key aspects.
% First, most rely strictly on monotonic decoding, meaning that once a token is emitted, it cannot be revised even if subsequent visual cues contradict the initial prediction.
% Second, they are predominantly developed on pre-segmented, sentence-level inputs, failing to address the challenges of \textit{unsegmented, long-form} streams (e.g., boundary detection) in real-world deployment.
\section{Method}

\begin{wrapfigure}{r}{0.5\textwidth}
\vspace{-1.8\baselineskip}
  \begin{center}
    \centerline{\includegraphics[width=0.5\textwidth]{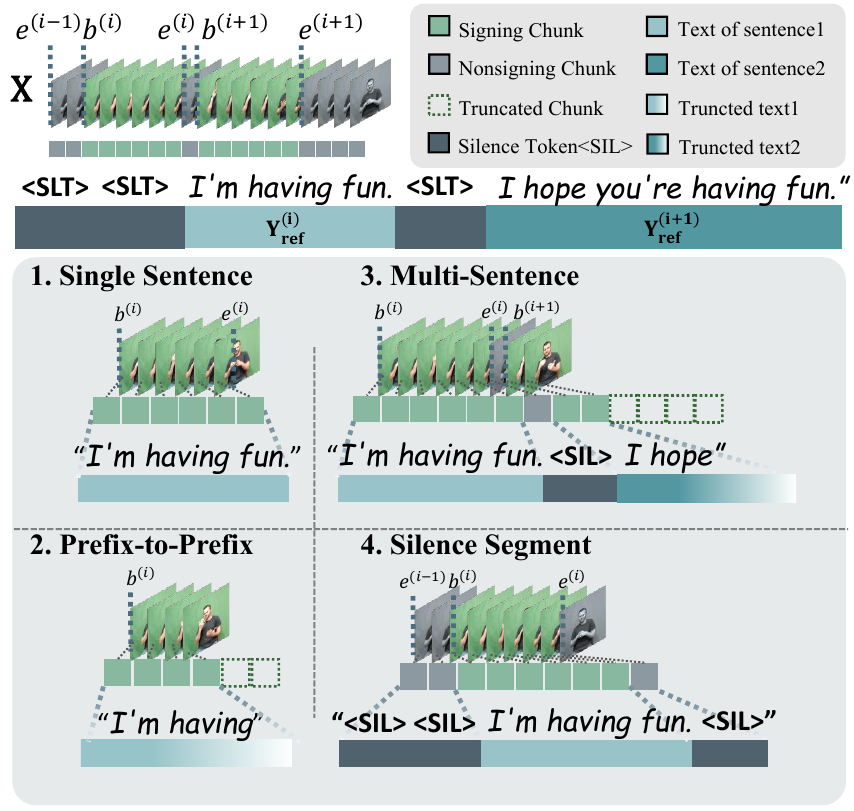}}
    \caption{
      Inference-aware training data construction.
From a long-form sign language video containing at least two sentences (top), besides the standard sentence-level clips, we construct three types of augmented training instances to reflect streaming inference conditions:
(1) prefix-to-prefix instances with truncated video and text,
(2) unsegmented multi-sentence clips,
and (3) clips augmented with surrounding non-signing segments represented by the \texttt{<SIL>} token.
    }
    \label{fig:inference_aware_ft}
  \end{center}
  \vspace{-7mm}
\end{wrapfigure}

\subsection{Overview}
As illustrated in Fig.~\ref{fig:overview}, we address Simultaneous Sign Language Translation by invoking a tailored SLT model $\mathcal{F}$ under a dedicated streaming inference scheme.
The model accepts an RGB video stream $\mathbf{X}$ of arbitrary length as input and produces the corresponding natural-language translation for each video segment. 
We adopt a similar Uni-Sign~\cite{li2025uni} architecture for our task. 
Specifically, the model consists of three components: a human pose estimator~\cite{jiang2024rtmwrealtimemultiperson2d}, a pose feature extraction network~\cite{Yan_Xiong_Lin_2018}, and a multimodal language model~\cite{xue-etal-2021-mt5} as text decoder. Please refer to the appendix~\ref{sec:appendix_slt_details} for more details.

Inspired by \emph{re-translation} paradigm in speech translation~\cite{ArivazhaganRetrans,arivazhagan2020retranslationversusstreamingsimultaneous}, we introduce an analogous scheme for video, operating on a chunk-by-chunk basis.
The input video stream $\mathbf{X}$ is partitioned into non-overlapping chunks, each consisting of $\Delta$ consecutive frames.
As chunks arrive, they are accumulated in a dynamic \emph{input buffer}, as illustrated on the left side of Fig.~\ref{fig:overview}.

With the \emph{sentence commitment} mechanism, the system translates one sentence at a time at runtime.
Let $i$ denote the index of the target natural-language sentence currently being translated, and let $s^{(i)}$ be the start chunk index of its corresponding video segment in the input stream.
At time step $t$, the newly arrived chunk $\mathbf{x}_t$ is appended to the input buffer denoted as:
\[
\mathbf{X}_{s^{(i)}:t} = (\mathbf{x}_{s^{(i)}}, \dots, \mathbf{x}_t).
\]
The model produces a revisable hypothesis for the current sentence:
\[
\mathbf{Y}_t = \mathcal{F}(\mathbf{X}_{s^{(i)}:t}) = (y_{t,1}, \dots, y_{t,L_t}), \quad y_{t,j} \in \mathcal{V}'.
\]
where $y_{t,j}$ is the $j$-th token in the hypothesis and $\mathcal{V}'$ denotes an augmented vocabulary as detailed in Section~\ref{sec:fine_tune}. 
As more visual context becomes available, $\mathbf{Y}_t$ may be revised.

The user-facing translation at step $t$ consists of the concatenation $\mathbf{Y}^{(<i)} \oplus \mathbf{Y}_t$,
where $\mathbf{Y}^{(<i)}$ denotes the translation committed for sentences $1$ to $i-1$ and remains fixed once produced.
When sentence completion is detected, the finalized translation $\mathbf{Y}^{(i)}$ is extracted from $\mathbf{Y}_t$
and concatenated with $\mathbf{Y}^{(<i)}$ to form $\mathbf{Y}^{(<i+1)}$,
after which the start index of the next sentence is advanced to $s^{(i+1)}$,
as detailed in Section~\ref{sec:sentence_commitment}.

\subsection{Inference-Aware Training}
\label{sec:fine_tune}

Simultaneous SLT over unsegmented long-form videos involves partial inputs, non-signing intervals, and ambiguous sentence boundaries. 
Models trained only on sentence-level clips tend to hallucinate under these conditions.
To align training with such inference, we construct a mixed training corpus from long-form videos with sentence-level annotations, as illustrated in Fig.~\ref{fig:inference_aware_ft}.

We assume a sign language video $\mathbf{X}=(\mathbf{x}_1,\dots,\mathbf{x}_T)$ is associated with a set of sentence-level annotations $\{(b^{(i)}, e^{(i)}, \mathbf{Y}^{(i)}_{\text{ref}})\}_{i=1}^N$, where $b^{(i)}$ and $e^{(i)}$ denote the begin and end chunk indices of the $i$-th sentence,
and $\mathbf{Y}^{(i)}_{\text{ref}}$ is its reference translation.
Chunks within annotated sentence boundaries correspond to signing activity,
while chunks outside these boundaries correspond to either unannotated signing content or non-signing silence,
where the latter is identified using a detection method described in Appendix~\ref{sec:vad}.

Starting from standard sentence-level clips paired with complete target sentences,
\[
\mathbf{X}_{\text{sample}}=(\mathbf{x}_{b^{(i)}},\dots,\mathbf{x}_{e^{(i)}}), \quad
\mathbf{Y}_{\text{sample}}=\mathbf{Y}^{(i)}_{\text{ref}}.
\] 
we enable the model to handle these situations by deriving three types of inference-aware training instances
for the unconventional cases described earlier.

\textbf{Prefix-to-Prefix.} 
To expose the model to partial inputs, we truncate the signing segment to its first $k$ chunks,
where $k$ is uniformly sampled such that $b^{(i)} \leq k < e^{(i)}$:
\[
\mathbf{X}_{\text{sample}}=(\mathbf{x}_{b^{(i)}},\dots,\mathbf{x}_{k}), \quad 
\mathbf{Y}_{\text{sample}} = \text{Trunc}(\mathbf{Y}^{(i)}_{\text{ref}}, r).
\]
where $\text{Trunc}(\cdot, r)$ truncates the target sentence to its prefix with ratio $r=(k-b^{(i)}+1)/(e^{(i)}-b^{(i)}+1)$, following prior work~\cite{niehues2018low,ArivazhaganRetrans}.
This augmentation encourages the model to generate stable partial hypotheses without prematurely predicting sentence termination.
\begin{figure*}[t]
  \centering
  \includegraphics[width=0.9\textwidth,height=0.2\textheight]{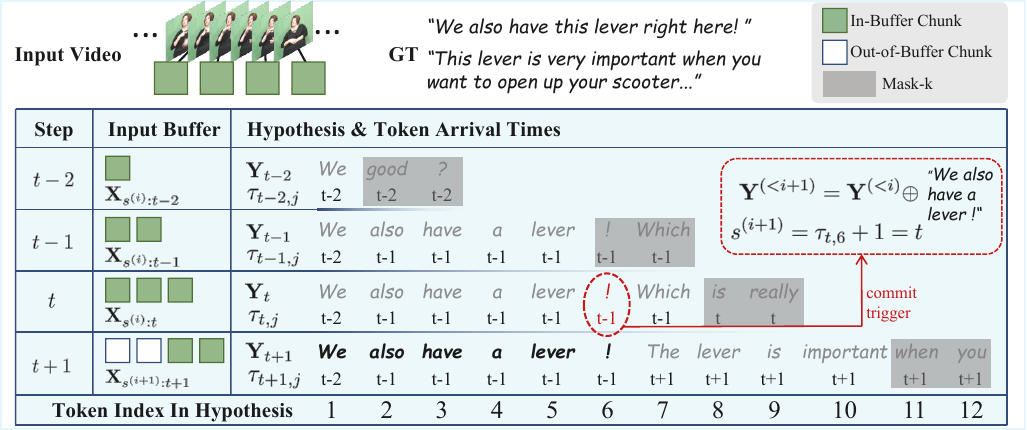}
  \caption{Stabilized re-translation with sentence commitment. At each step $t$, the model re-translates the buffered video input to produce a raw hypothesis $\tilde{\mathbf{Y}}_t$, from which a stabilized hypothesis $\mathbf{Y}_t$ is obtained by mask-$k$ (mask-$2$). Re-translation corrects premature predictions, including early terminal punctuation, as more context becomes available. When a terminal punctuation (e.g. ``!") remains stable in $\mathbf{Y}_t$, it triggers sentence commitment: all preceding tokens are finalized as the translation of the current sentence, appended to the history, and the input buffer is advanced based on the punctuation’s token arrival time.}
  \label{fig:inference}
  \vspace{-3mm}
\end{figure*}

\textbf{Silence Segments.}
Unlike acoustic silence in speech translation, non-signing intervals in SLT may contain large non-linguistic movements like dancing. 
Thus cannot be reliably detected by VAD-like tools during online inference.
We address this by integrating silence prediction into the SLT model, allowing it to identify non-signing content based on its understanding of sign language.

We augment the target vocabulary with a silence token \texttt{<SIL>}, denoted by the augmented vocabulary $\mathcal{V}'$.
Let $\mathcal{S}\subset\{1,\dots, T\}$ denote the set of non-signing chunk indices.
For each sentence, we optionally extend its boundaries by $\delta_b$ and $\delta_e$ chunks,
provided the extended region lies in $\mathcal{S}$ and does not overlap with adjacent sentences.
The resulting instance is
\[
\mathbf{X}_{\text{sample}} = (\mathbf{x}_{b^{(i)}-\delta_b}, \dots, \mathbf{x}_{e^{(i)}+\delta_e}),
\]
\[
\mathbf{Y}_{\text{sample}} = \texttt{<SIL>}^{\delta_b} \oplus \mathbf{Y}^{(i)}_{\text{ref}} \oplus \texttt{<SIL>}^{\delta_e}.
\]
Here, $\texttt{<SIL>}^{\delta}$ denotes $\delta$ consecutive silence tokens.
We additionally sample silence-only segments paired with a single \texttt{<SIL>} token to suppress hallucination during extended pauses.

\textbf{Multi-sentence.} 
To reflect cases where the input buffer spans multiple sentences, temporally adjacent sentences are treated as a single training instance when the temporal gap between them is less than two seconds, a conservative criterion to avoid including unannotated signing content between sentences.
\[
\mathbf{X}_{\text{sample}}=(\mathbf{x}_{b^{(i)}},\dots,\mathbf{x}_{e^{(i+1)}}), \quad
\mathbf{Y}_{\text{sample}}=\mathbf{Y}^{(i)}_{\text{ref}} \oplus \mathbf{Y}^{(i+1)}_{\text{ref}},
\]
If silence chunks occur between the two sentences, the corresponding number of \texttt{<SIL>} tokens are inserted into $\mathbf{Y}_{\text{sample}}$.

During training, these instance types are dynamically mixed.
Prefix truncation, silence augmentation, and multi-sentence concatenation
are applied with probabilities $p_{\text{prefix}}$, $p_{\text{silence}}$, and $p_{\text{multi}}$, respectively.
Implementation details are provided in the Appendix~\ref{sec:appendix_training_pipeline}.

\subsection{Stabilized Re-Translation}
Re-translation generates a hypothesis from scratch at each step.
As visual input changes over time, token probability shifts in the decoder language model lead to radical rewrites of previously generated text, causing visual flicker.
To mitigate this issue, we adopt two stabilization techniques, biased search and output masking~\cite{ArivazhaganRetrans,arivazhagan2020retranslationversusstreamingsimultaneous}.

Given the buffered input $\mathbf{X}_{s^{(i)}:t}$, the model first produces a raw hypothesis $\tilde{\mathbf{Y}}_t = (\tilde{y}_{t,1}, \dots, \tilde{y}_{t,\tilde{L}_t})$,
which is then refined to produce the stabilized output $\mathbf{Y}_t = (y_{t,1}, \dots, y_{t,L_t})$.

\textbf{Biased Search.}
To promote temporal consistency, we generate the raw hypothesis $\tilde{\mathbf{Y}}_t$ using beam search with a bias toward the previous stabilized output $\mathbf{Y}_{t-1}$.
When predicting the token $\tilde{y}_{t,j}$ at position $j$, the probability of each candidate token $w \in \mathcal{V}'$ is adjusted as:
\[
P_{\text{biased}}(\tilde{y}_{t,j} = w) = (1 - \beta) P(\tilde{y}_{t,j} = w) + \beta \cdot \mathbb{I}(y_{t-1,j} = w),
\]
where $P(\cdot)$ denotes the original language model distribution, $\beta \in [0,1]$ controls the bias strength, and $\mathbb{I}(\cdot)$ is the indicator function.
The bias is applied only when the current prefix up to position $j\!-\!1$
matches that of the previous stabilized hypothesis,
i.e., $(\tilde{y}_{t,1}, \dots, \tilde{y}_{t,j-1}) = (y_{t-1,1}, \dots, y_{t-1,j-1})$,
and no sentence commitment occurred at the previous step.
Once the model deviates from the previous output, the bias is disabled to allow necessary corrections.

\textbf{Mask-$k$.}
To suppress unstable suffix tokens caused by insufficient visual context, we apply a masking operation to the raw hypothesis. 
Given $\tilde{\mathbf{Y}}_t$ of length $\tilde{L}_t$, the stabilized output is obtained as
\[
\mathbf{Y}_t = \text{Mask}(\tilde{\mathbf{Y}}_t, k) = (y_{t,1}, \dots, y_{t,L_t}),
\]
where $L_t = \max(0, \tilde{L}_t - k)$ and $y_{t,j} = \tilde{y}_{t,j}$ for $j \le L_t$.
Mask is applied only when the input video is incomplete; the final translation obtained after observing the entire video is generated without masking.

\subsection{Sentence Commitment}
\label{sec:sentence_commitment}

Since the input is an unsegmented video stream of arbitrary length, corresponding to an arbitrary number of sentences, accumulating the entire video is infeasible.
To bound memory usage and segment the continuous input stream at runtime, we adopt a sentence commitment mechanism, illustrated in Fig.~\ref{fig:inference}. 
Rather than introducing special boundary tokens~\cite{iranzo-sanchez-etal-2024-segmentation}, ReSCUE uses terminal punctuation as the commitment signal, leveraging the language model's semantic understanding of sentence completion.

A commitment is triggered if the stabilized hypothesis $\mathbf{Y}_t$ contains a terminal punctuation token (e.g., “.”, “!”, “?”) or consists solely of the token $\texttt{<SIL>}$.
Let $j^*$ denote the position of the \emph{first} terminal punctuation token in $\mathbf{Y}_t$.
Upon commitment, the finalized translation of the current sentence is extracted as $\mathbf{Y}^{(i)} = (y_{t,1},\dots,y_{t,j^*})$, which is then appended to the committed history:
\[
\mathbf{Y}^{(<i+1)} = \mathbf{Y}^{(<i)} \oplus \mathbf{Y}^{(i)},
\]
where $\mathbf{Y}^{(<1)}$ is initialized as an empty sequence.
The input buffer is then advanced to discard visual content associated with the finalized sentence.
To determine a precise cut point in the video stream, we associate each token $\tilde{y}_{t,j}$ in the raw hypothesis $\tilde{\mathbf{Y}}_t$
with an arrival time $\tau_{t,j}$, defined as the chunk index at which the token at position $j$ was first generated and remains unchanged thereafter.
Arrival times are updated recursively:
\[
\tau_{t,j} =
\begin{cases}
\tau_{t-1,j}, & \text{if } \tilde{y}_{t,j} = \tilde{y}_{t-1,j}, \\
t, & \text{otherwise}.
\end{cases}
\]
Newly generated tokens at step $t$ are assigned an arrival time $t$.
This definition ensures arrival times are unaffected by mask-$k$ and captures the earliest point at which sufficient visual evidence supports token generation.

If a commitment is triggered by a terminal punctuation token at position $j^*$, the start index of the next sentence is set to $s^{(i+1)} = \tau_{t,j^*}+1$. 
If the stabilized hypothesis consists of a single $\texttt{<SIL>}$ token, we set $s^{(i+1)} = t+1$.
By advancing the buffer according to token arrival times rather than the current step, we avoid discarding visual evidence relevant to subsequent sentences.
\section{Long-Form Dataset}
\label{sec:dataset}
We introduce \textbf{How2Sign-Long} and \textbf{CSL-Story} for both model training and realistic evaluation. Full documentation (segmentation, statistics, and samples) is deferred to Appendix~\ref{sec:long-form dataset}.

\noindent

\begin{wraptable}[24]{r}{0.42\textwidth}
\vspace{-1\baselineskip}
\centering
\caption{Comparison on \textit{CSL-Daily}. \textbf{Bold} values indicate the best offline SLT result; \underline{underline} indicates the best simultaneous SLT result. \same~denotes the same AL as SLRT; for offline SLT models, AL is determined by the average duration of the full test samples. Zuo et al.~\cite{zuo2024towards} is excluded from latency comparison since its reported latency is not end-to-end video-to-text.}
\vspace{-0.8em}
\footnotesize
\setlength{\tabcolsep}{2.3pt}
\renewcommand{\arraystretch}{0.86}
\begin{tabular}{@{}lccc@{}}
\toprule
Method & AL(ms)$\downarrow$ & BLEU$\uparrow$ & ROUGE$\uparrow$ \\
\midrule
\rowcolor{gray!15}\multicolumn{4}{@{}c@{}}{\textbf{offline SLT}} \\
SLRT~\cite{Camgoz_2020_CVPR}      & 4259  & 3.03 & 19.67 \\
GASLT~\cite{yin2023gloss}         & \same & 4.07 & 20.35 \\
GFSLT-VLP~\cite{zhou2023gloss}    & \same & 11.00 & 36.44 \\
FLa-LLM~\cite{chen2024factorized} & \same & 14.20 & 37.25 \\
Sign2GPT~\cite{wong2024sign2gpt}  & \same & 15.40 & 42.36 \\
SignLLM~\cite{gong2024llms}       & \same & 15.75 & 39.91 \\
SignCL~\cite{ye2024improving}     & \same & 16.16 & 48.92 \\
MSLU~\cite{zhou2025scaling}       & \same & 11.42 & 33.80 \\
C$^2$RL~\cite{chen2025c}          & \same & 21.61 & 48.21 \\
CV-SLT~\cite{zhao2024conditional} & \same & \textbf{28.94} & \textbf{57.06} \\
Uni-Sign~\cite{li2025uni}         & \same & 25.61 & 54.92 \\
\midrule
\rowcolor{gray!15}\multicolumn{4}{@{}c@{}}{\textbf{simultaneous SLT}} \\
SimulSLT ($k$=7)~\cite{simulslt}  & 3370 & 13.88 & 43.18 \\
SimulSLT ($k$=1)                  & 1427 & 12.30 & 40.50 \\
CTL++ ($m$=6)~\cite{sun-etal-2024-adaptive} & 2427 & 14.46 & 41.68 \\
CTL++ ($m$=0)                     & 1163 & 14.13 & 41.62 \\
Zuo et al.~\cite{zuo2024towards}  & --   & 20.63 & -- \\
\midrule
\textbf{ReSCUE} ($k$=0)           & \underline{693} & 23.32 & 51.67 \\
\textbf{ReSCUE} ($k$=6)           & 2438 & \underline{24.72} & \underline{54.02} \\
\bottomrule
\end{tabular}
\vspace{-0.5em}
\begin{flushleft}
\scriptsize
\end{flushleft}
\vspace{-1.0\baselineskip}
\label{tab:slt_main_csl}
\end{wraptable}

\textbf{How2Sign-Long.}  
% We construct How2Sign-Long by re-segmenting the raw footage from which the original How2Sign~\cite{DBLP:journals/corr/abs-2008-08143} clips were derived. 
% Unlike the standard dataset's single-sentence clips, our samples contain continuous multiple sentences and natural non-signing segments. 
% The precise temporal boundaries preserved for each constituent sentence facilitate Inference-Aware Training data construction.
% For fair comparison with prior work, we adopt the official data splits and preserve the original sentence-level train/test distribution.
The original How2Sign dataset contains long videos, but some signed segments do not have text annotations, making direct quantitative evaluation impossible.
We therefore reconstruct How2Sign-Long by reassembling temporally continuous and fully annotated sentence segments from the original videos.
The reconstructed samples contain multiple sentences and natural non-signing segments.
For fair comparison with prior work, we follow the official data splits and preserve the original sentence-level train/test distribution.

% This construction results in long-form video samples with an average duration of \textbf{[XX–YY seconds]}, containing \textbf{[N]} sentences per clip on average.
% The dataset contains \textbf{[M]} training samples and \textbf{[K]} long-form evaluation samples.
% 特性2：更重要的是，原版How2Sign的分割方式，N long-form，N1 + N2 = N ->我们在Long 上面的train/test数据，会和原来的How2sign高度重合/一致，更方便和已有的在How2Sign上测试过的工作进行比较

\noindent

\textbf{CSL-Story.} 
Due to the lack of continuous Chinese Sign Language (CSL) video data, we curated CSL-Story, a dataset derived from fairy tale narrations on a publicly accessible social media platform.
Chinese subtitles were extracted via  Optical Character Recognition (OCR)~\cite{umiocr} and aligned with the videos, providing accurate sentence-level temporal boundaries for each sample.

% Compared to How2Sign-Long, CSL-Story is relatively small, containing \textbf{[A]} videos with an average duration of \textbf{[B]} seconds. We use this dataset solely for evaluation to verify that our approach remains effective and robust on real-world, naturally occurring CSL long-form streams.

\section{Experiments}
To evaluate our method's translation quality, latency, and stability, we conduct experiments on both sentence-level and long-form sign language datasets. We provide the demonstration video in the supplementary material to further illustrate our method's performance.

\subsection{Experiment Setup}

\textbf{Datasets.}
We evaluate our method under two settings: \textbf{sentence-level}, where each video corresponds to a single target sentence, and \textbf{long-form}, where each video consists of multiple unsegmented, contiguous sentences. For sentence-level evaluation, we follow prior work \cite{simulslt, sun-etal-2024-adaptive, zuo2024towards} and conduct experiments on CSL-Daily \cite{zhou2021improving} and Phoenix-2014-T \cite{camgoz2018neural}. We further evaluate on two long-form datasets, CSL-Story and How2Sign-Long. 

\textbf{Evaluation Metrics.}
We evaluate SLT quality using BLEU~\cite{papineni2002bleu} (computed with SacreBLEU~\cite{post2018call}) and ROUGE-L \cite{lin2004rouge}. Latency is measured by Average Lagging (AL)~\cite{ma2019stacl} and Length-Adaptive Average Lagging (LAAL)~\cite{Papi_2022}. Following \cite{arivazhagan2020retranslationversusstreamingsimultaneous}, the AL is calculated based on token stabilization time. 
To assess translation stability, we adopt Normalized Erasure (NE) \cite{arivazhagan2020retranslationversusstreamingsimultaneous}, a metric widely utilized in simultaneous speech translation, which quantifies output ``flickering'' by calculating the ratio of the total number of deleted tokens to the final sequence length.

\textbf{Implementation Details.} 
Our model is trained using four NVIDIA H20 GPUs, with a per-GPU batch size of 8. We optimize the model using AdamW with an initial learning rate of $1\times10^{-4}$ and a weight decay of $1\times10^{-4}$. 
A cosine decay learning rate scheduler is applied throughout training. 
Training and inference are conducted with a chunk size $\Delta=15$, and a biased search weight $\beta=0.2$. 
We further test inference speed on a single NVIDIA RTX 3090 GPU, showing that ReSCUE can run in real time and is deployable in practical settings, as detailed in Appendix~\ref{sec:appendix_inference_speed}.

\textbf{Training Strategies.}
All datasets incorporate Prefix-to-Prefix instances ($p_{\text{prefix}}=0.5$). For sentence-level datasets, $p_{\text{multi}}$ and $p_{\text{silence}}$ are fixed at $0$ due to dataset constraints. Conversely, training on long-form datasets includes Silence Segments ($p_{\text{silence}}=0.2$) and Multi-sentence instances ($p_{\text{multi}}=0.5$).

\subsection{Sentence-level SLT Performance}

\begin{wrapfigure}[19]{r}{0.42\textwidth}
\vspace{-1.8\baselineskip}
  \centering
    \includegraphics[width=0.8\textwidth, height=0.15\textheight, keepaspectratio]{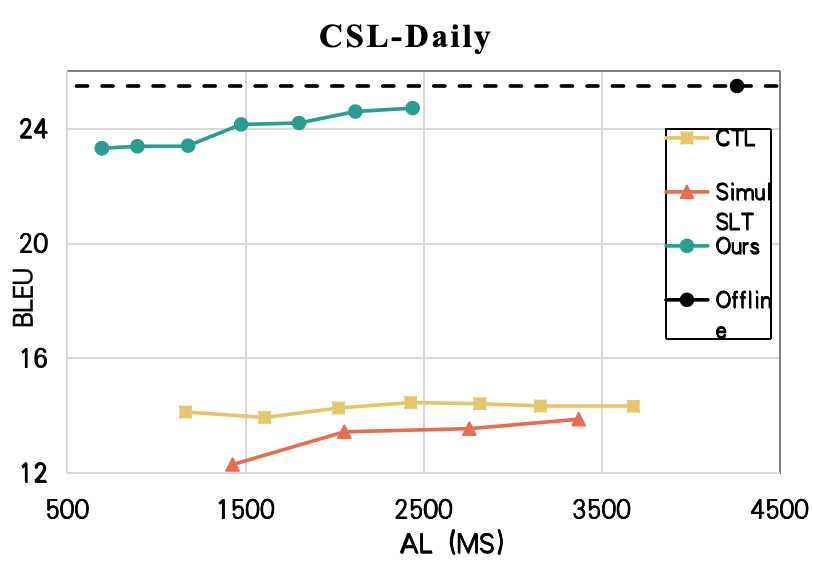}
  \caption{Sentence-Level Translation quality (BLEU) vs. latency (Average Lagging) on CSL-Daily. For CTL++, latency is controlled by parameter $m$, with points from left to right corresponding to $m \in \{0, 2, 4, 6, 8, 10, 18\}$. For SimulSLT, latency is adjusted via the wait-$k$ parameter $k_{wait}$, with points from left to right representing $k_{wait} \in \{1, 3, 5, 7\}$. For ReSCUE, latency is adjusted via the Mask-$k$ parameter $k$, with points from left to right representing $k \in \{0, 1, 2, 3, 4, 5, 6\}$. The \textbf{top-left} region represents the optimal performance.
}
  \label{fig:al_bleu4_csldaily}
\end{wrapfigure}

\textbf{Comparison with Simultaneous SLT.} 
We compare ReSCUE with representative simultaneous sign language translation methods.
\textbf{SimulSLT} \cite{simulslt} employs a \textit{wait-k} strategy that initiates translation only after observing $k$ glosses. 
More recently, \textbf{CTL++} \cite{sun-etal-2024-adaptive} introduces latency-aware modeling and enhances online decoding by removing $m$ attempted tokens at each step.

Compared with prior simultaneous SLT methods, ReSCUE achieves lower latency while maintaining strong translation quality in low-latency regimes, where competing methods degrade substantially as latency decreases (Fig.~\ref{fig:al_bleu4_csldaily}).
On CSL-Daily (Table~\ref{tab:slt_main_csl}), our fastest configuration ($k=0$) achieves a \textbf{40.4\% reduction in latency} (693~ms vs. 1163~ms) compared to the strongest simultaneous baseline, while our high-accuracy variant ($k=6$) improves \textbf{BLEU by 19.8\%} and \textbf{ROUGE by 29.6\%} over existing simultaneous works \cite{zuo2024towards, sun-etal-2024-adaptive}. 
The robustness of ReSCUE is further supported by supplementary experiments on Phoenix-2014-T (Fig.~\ref{fig:al_bleu4_phx} and Table~\ref{tab:complete_phx}).

\textbf{Comparison with Offline SLT.} 
We compare ReSCUE with offline SLT models to evaluate the performance gap.
For every sentence-level dataset, we report two boundary configurations: one for \textbf{minimum latency} and one for \textbf{maximum quality}. 
On CSL-Daily, the high-quality configuration ($k=6$) remains close to the state-of-the-art offline model CV-SLT~\cite{zhao2024conditional}, with gaps of only 4.22 BLEU and 3.04 ROUGE, while operating simultaneously.
Additional results on Phoenix-2014-T are reported in the appendix (Table~\ref{tab:complete_phx}).

While offline SLT models leverage global context for higher quality, their requirement for full-video input results in latencies exceeding 4,000~ms. 
Conversely, ReSCUE maintains comparable translation quality at significantly lower latency (693~ms).

\subsection{Long-form SLT Performance}

Since existing simultaneous SLT methods are limited to sentence-level data, direct comparisons on long-form data are currently not possible. 
As shown in Table~\ref{tab:results_longform}, directly applying the offline SLT model Uni-Sign~\cite{li2025uni} to long-form videos leads to significant degradation.

For a more meaningful reference, we report an \textbf{Oracle Reference} (O-R), where ground-truth sentence boundaries are provided to Uni-Sign so that each long-form video is translated as a sequence of single-sentence clips.
The resulting sentence-level translations are then concatenated to form the final output.
Although such oracle boundaries are unavailable during practical inference, this setting provides a strong reference for long-form translation quality.
ReSCUE achieves comparable quality on both How2Sign-Long and CSL-Story, while reducing latency from tens of seconds to about 1.6 seconds.
These results show that ReSCUE can jointly handle online boundary detection and translation on unsegmented long-form videos.

We also report results on sentence-level How2Sign in the appendix (Table~\ref{tab:complete_how2sign}), showing that ReSCUE is effective across both sentence-level and long-form settings.
We provide qualitative long-form translation examples in Appendix~\ref{sec:failure_cases}, including successful cases and representative failure modes.

\begin{table}[t]
\centering
\footnotesize

\begin{minipage}[t]{0.49\linewidth}
\centering
\captionof{table}{Translation performance on long-form datasets. \textbf{Bold} indicates the best result. O-R denotes an oracle reference using ground-truth sentence boundaries, which is unattainable in practical inference.}
\label{tab:results_longform}
\vspace{0.4\baselineskip}
\setlength{\tabcolsep}{5pt}
\renewcommand{\arraystretch}{0.92}
\begin{tabular}{@{}lccc@{}}
\toprule
\textbf{Method} & AL$\downarrow$ & BLEU$\uparrow$ & ROUGE$\uparrow$ \\
\midrule
\rowcolor{gray!15}\multicolumn{4}{c}{\textbf{CSL-Story}} \\
Uni-Sign & 37722 & 2.10 & 32.42 \\
Uni-Sign (O-R) & 37722 & \textbf{24.80} & \textbf{46.71} \\
\textbf{ReSCUE} ($\beta$=0.2, $k$=3) & \textbf{1588} & 23.64 & 41.38 \\
\midrule
\rowcolor{gray!15}\multicolumn{4}{c}{\textbf{How2Sign-Long}} \\
Uni-Sign & 53687 & 0.51 & 24.16 \\
Uni-Sign (O-R) & 53687 & \textbf{15.59} & \textbf{39.61} \\
\textbf{ReSCUE} ($\beta$=0.2, $k$=1) & \textbf{1580} & 14.71 & 38.42 \\
\bottomrule
\end{tabular}
\end{minipage}
\hfill
\begin{minipage}[t]{0.49\linewidth}
\centering
\captionof{table}{Ablation study on CSL-Story.  \textbf{Bold} indicates the best result. We report both AL and LAAL because AL can be misleading when models over-generate or commit prematurely. }
\label{tab:exp_long_form_ablation}
\vspace{0.4\baselineskip}
\setlength{\tabcolsep}{2pt}
\renewcommand{\arraystretch}{0.92}
\begin{tabular}{@{}lcccc@{}}
\toprule
\textbf{Method} & AL$\downarrow$ & LAAL$\downarrow$ & BLEU$\uparrow$ & NE$\downarrow$ \\
\midrule
\rowcolor{gray!15}\multicolumn{5}{c}{\textbf{CSL-Story}} \\
\midrule
\textbf{ReSCUE} & \textbf{1588} & 2725 & \textbf{23.64} & \textbf{0.15} \\
\midrule
w/o Sentence Commit & 10093 & 10431 & 7.91 & 1.01 \\
w/o Silence Segments & -473 & 1699 & 19.26 & 0.32 \\
w/o Prefix-to-Prefix & -2470 & \textbf{1674} & 21.89 & 0.31 \\
w/o Multi-sentence & 13525 & 13663 & 9.85 & 0.76 \\
\bottomrule
\end{tabular}
\end{minipage}

\end{table}

\subsection{Ablation Studies}

\textbf{Impact of Inference-aware Training.}
Table~\ref{tab:exp_long_form_ablation} evaluates the contribution of each inference-aware training component.
Removing any of the training strategies lowers BLEU, showing that partial inputs, non-signing intervals, and multi-sentence contexts are all necessary for long-form streaming SLT.
The largest degradation comes from removing \textbf{Multi-sentence} training, which leads to much higher latency and unstable outputs.
Removing \textbf{Silence Segments} or \textbf{Prefix-to-Prefix} training also increases NE and produces abnormally low, sometimes negative AL, caused by hallucinated outputs or premature commitments.
Both failures can make the prediction much longer than the ground truth, which artificially lowers the computed latency.
We discuss these failure modes in Section~\ref{sec:limitations_discussion}.

\begin{wrapfigure}[22]{r}{0.5\textwidth}
\vspace{-1.7\baselineskip}
    \centering
    \includegraphics[width=\linewidth]{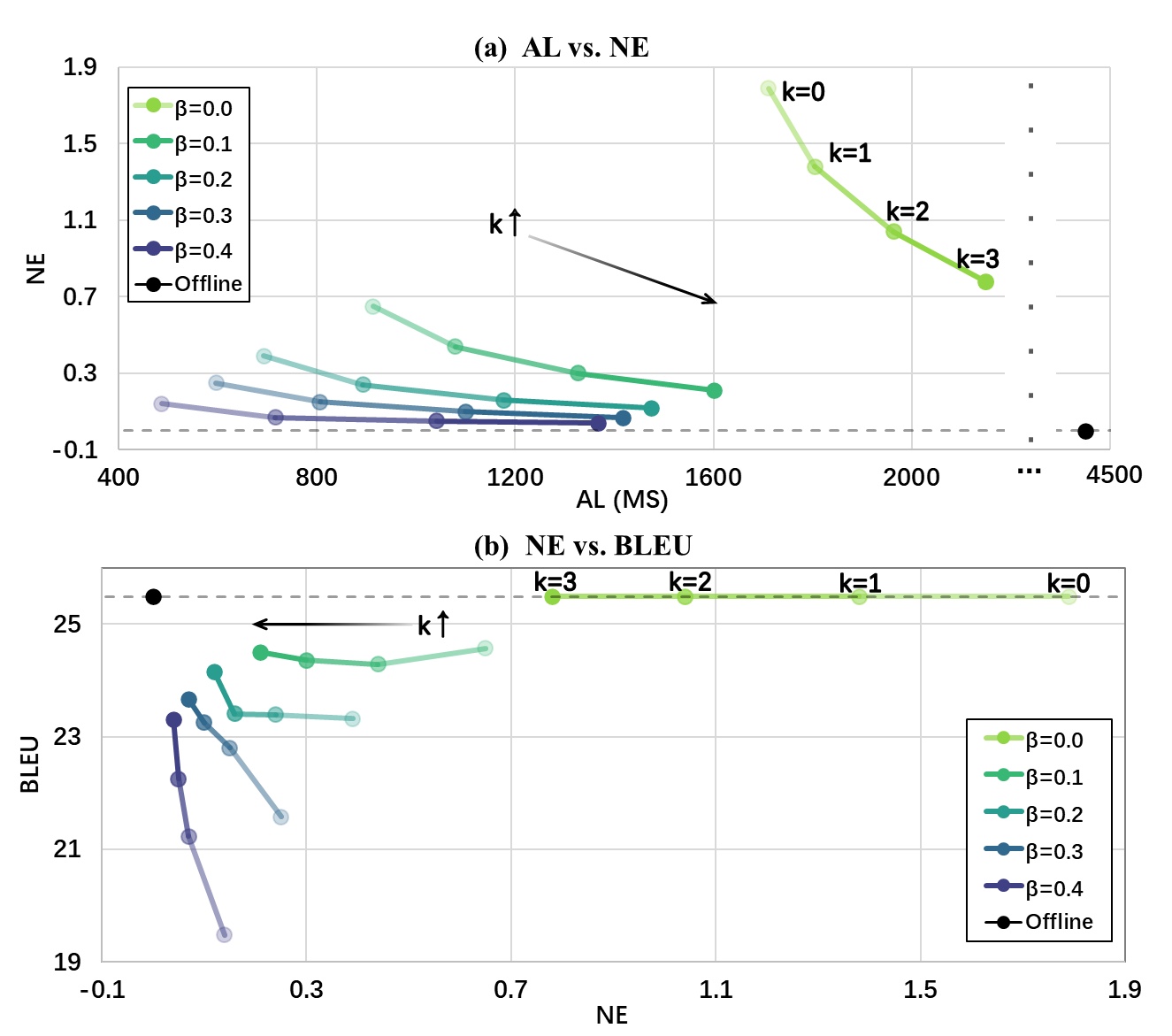}
    \label{fig:abl_al_ne_csl}
%\vspace{0.1in}
  \caption{Ablation study on sentence-level Simultaneous SLT. We explore the trade-offs between latency (Average Lagging), stability (NE), and translation quality (BLEU). $\beta$ and $k$ refer to the bias strength in biased search and the parameter for the Mask-$k$ strategy, respectively. In subfigure (a), the bottom-left region represents the optimal performance, while in subfigure (b), the top-left region represents the optimal performance.}
  \label{fig:ablaiton_stability_ctrl}
  \vspace{-9mm}
\end{wrapfigure}

\textbf{Impact of Stabilized Re-Translation.}
As shown in Fig.~\ref{fig:ablaiton_stability_ctrl}, the biased search parameter $\beta$ and mask-$k$ parameter $k$ jointly regulate the trade-off between latency, quality, and stability. 

Increasing $\beta$ reduces Normalized Erasure (NE), improving translation stability, but can reduce BLEU when revisions are overly constrained.
For a fixed $\beta$, larger $k$ suppresses unstable suffix tokens and further lowers NE, at the cost of higher AL. 
Notably, in configurations where $\beta > 0.2$, increasing $k$ improves BLEU. 
This suggests that mask-$k$ prevents unstable trailing tokens from being fed into subsequent biased search steps, thereby mitigating error propagation. 
Based on these results, we set $\beta=0.2$ and choose $k$ according to the target latency--quality trade-off.

\textbf{Impact of Sentence Commitment.} As shown in Table~\ref{tab:exp_long_form_ablation}, without sentence commitment degrades BLEU and AL/LAAL a lot. Without it, the input buffer expands indefinitely, causing the model to treat the entire stream as a single sentence. Additional results are provided in the appendix (Table~\ref{tab:complete_long_form_ablation}).

\section{Limitations \& Discussion}
\label{sec:limitations_discussion}

Although ReSCUE approaches oracle-reference quality, simultaneous long-form SLT introduces failure modes absent from standard sentence-level translation.

\textbf{Premature commitment.}
The model may predict terminal punctuation before sufficient visual context is observed, causing
irreversible premature commitment and making the next sentence start from the middle of the true
signing segment.
This is severe without Prefix-to-Prefix training, where the model tends to generate complete sentences with terminal punctuation even from short prefixes.
Mask-$k$ helps mitigate this problem by hiding unstable early punctuation, giving the model a chance to revise it when more visual evidence arrives, as shown in Table~\ref{abl_longform_bk}.

\textbf{Delayed commitment.}
The model may also postpone terminal punctuation for too long, causing the input buffer to keep growing,
earlier correct translations to be overwritten, and multiple sentences to be treated as one.
This is more frequent without Multi-sentence training, where the model is not exposed to sentence transitions inside the input buffer.
As a result, terminal punctuation may only appear near the output end and be repeatedly hidden by Mask-$k$.

\textbf{Silence detection errors.}
Although less frequent, incorrect \texttt{<SIL>} predictions can discard useful signing frames, especially near the beginning of a sentence, leading to missing subjects or incomplete translations.

% ReSCUE also inherits common SLT errors such as noisy pose input and oov signs, and the current evaluation is limited coverage across sign languages, signing styles, signer populations. We provide representative examples (Appendix~\ref{sec:failure_cases}) and quantitative evaluation of commitment (Appendix~\ref{quantitative_commitment}).
% We discuss broader societal impacts, potential risks, and responsible-use considerations in Appendix~\ref{sec:broader_impact}.

ReSCUE may also inherit common SLT errors such as noisy pose estimation and out-of-vocabulary signs.
We provide representative examples (Appendix~\ref{sec:failure_cases}) and quantitative evaluation of commitment (Appendix~\ref{quantitative_commitment}). We discuss broader impacts, potential risks, and responsible-use considerations in Appendix~\ref{sec:broader_impact}.
\section{User Study}

% We recruited 4 Deaf and 8 hearing volunteers to participate in the user study to evaluate the practical effectiveness of our simultaneous SLT system in real-world scenarios, comparing it directly against two different offline methods. ~\cite{li2025uni,zhao2024conditional}
% All participants provided informed consent, with instructions communicated in Sign Language for Deaf volunteers, and were compensated for their participation.
% % ---------------------------------

% The study consists of two complementary parts designed to assess both interactive communication and long-form translation capabilities.
% We collected 120 sign language videos performed by Deaf signers.
% The translations from the three methods were presented to hearing participants, yielding 960 subjective ratings and 480 sets of objective metrics. 
% Detailed experimental setups and metric definitions are provided in Appendix~\ref{sec:user_study}. 

% Across both studies, ReSCUE achieves higher subjective ratings in quality and fluency (Table~\ref{tab:user_study}(a)), as well as high accuracy and faster response times in the long-form task (Table~\ref{tab:user_study}(b)). These results demonstrate the effectiveness of real-world sign language communication.
To evaluate the practical effectiveness of our simultaneous SLT system in real-world scenarios, we conduct a user study comparing ReSCUE with two offline SLT methods. 
ReSCUE achieves better subjective quality and fluency, improving quality by 22.9\% and fluency by 37.5\% compared with the strongest baseline. 
ReSCUE achieves the fastest response speed among all methods in real-time translation scenarios while maintaining competitive selection accuracy. 
These results demonstrate the practical effectiveness of ReSCUE for real-world sign language communication. 
Detailed experimental setups, metric definitions, results, as well as participant instructions, written informed consent procedure, compensation policy, and review considerations are provided in Appendix~\ref{sec:user_study}.

\section{Conclusion}

We presented ReSCUE, an approach for Simultaneous SLT that operates on unsegmented long-form sign language videos. By aligning training with realistic streaming conditions, adapting re-translation to SLT, and utilizing sentence commitment for online segmentation, ReSCUE achieves low-latency and stable translation under continuous video input.

Experiments on both sentence-level and long-form settings demonstrate state-of-the-art performance for simultaneous SLT across diverse latency constraints and validate the effectiveness of ReSCUE in realistic streaming scenarios. We hope this work helps move Simultaneous SLT beyond sentence-level settings toward practical deployment on long-form video streams.

% \section*{References}
\clearpage
\bibliography{ref}
\bibliographystyle{IEEEtran}

% References follow the acknowledgments in the camera-ready paper. Use unnumbered first-level heading for
% the references. Any choice of citation style is acceptable as long as you are
% consistent. It is permissible to reduce the font size to \verb+small+ (9 point)
% when listing the references.
% Note that the Reference section does not count towards the page limit.
% \medskip

% {
% \small

% [1] Alexander, J.A.\ \& Mozer, M.C.\ (1995) Template-based algorithms for
% connectionist rule extraction. In G.\ Tesauro, D.S.\ Touretzky and T.K.\ Leen
% (eds.), {\it Advances in Neural Information Processing Systems 7},
% pp.\ 609--616. Cambridge, MA: MIT Press.

% [2] Bower, J.M.\ \& Beeman, D.\ (1995) {\it The Book of GENESIS: Exploring
%   Realistic Neural Models with the GEneral NEural SImulation System.}  New York:
% TELOS/Springer--Verlag.

% [3] Hasselmo, M.E., Schnell, E.\ \& Barkai, E.\ (1995) Dynamics of learning and
% recall at excitatory recurrent synapses and cholinergic modulation in rat
% hippocampal region CA3. {\it Journal of Neuroscience} {\bf 15}(7):5249-5262.
% }

%%%%%%%%%%%%%%%%%%%%%%%%%%%%%%%%%%%%%%%%%%%%%%%%%%%%%%%%%%%%

\appendix
\clearpage
\begin{center}
    \rule{0.99\textwidth}{0.8pt} \\[0.2em]
    \LARGE \textbf{Appendix} \\
    \rule{0.99\textwidth}{0.8pt}
\end{center}
\section{Complete Experiment}

\subsection{Comprehensive Sentence-level SLT performance comparison}
We present the complete performance on the CSL-Daily, Phoenix-2014-T, and How2Sign datasets in Tables~\ref{tab:complete_daily}, \ref{tab:complete_phx}, and \ref{tab:complete_how2sign}, respectively. 
To further evaluate the robustness of ReSCUE under varying latency constraints, we extend the latency-quality trade-off analysis to the Phoenix-2014-T dataset. As illustrated in Fig.~\ref{fig:al_bleu4_phx}, ReSCUE has lower latency, and achieves higher translation quality in the low latency regime compared to other methods.

\subsection{Comprehensive SLT Performance on Long-form Datasets}
We report comprehensive results on the CSL-Story and How2Sign-Long datasets in Table~\ref{tab:complete_long_form}. These results further demonstrate that our method closely approaches the Oracle Reference, which represents a theoretical upper bound unattainable in practice, highlighting our system’s strong performance in realistic sign language translation scenarios.

% \begin{table}[t]
% \centering
% \caption{Comprehensive SLT performance on long-form dataset}
% \begin{tabular}{l cccccccc}
% \toprule
% \textbf{Model} & AL$\downarrow$ & LAAL$\downarrow$ & BLEU1$\uparrow$ &BLEU2$\uparrow$ &BLEU3$\uparrow$ &BLEU4$\uparrow$ &Rouge$\uparrow$ & NE$\downarrow$ \\
% \midrule
% \rowcolor{gray!15}\multicolumn{9}{c}{\textbf{CSL-Story}} \\
% Uni-Sign       & 37722 & 37722 & 4.22 &3.31 &2.67 & 2.10 & 32.43 & 0.00 \\
% Uni-Sign(Oracle reference)       & 37722 & 37722 & 4.22 &3.31 &2.67 & 2.10 & 32.43 & 0.00 \\
% \textbf{Our}($\beta$=0.2, $k$=3)        & 1588 & 2725 & 54.48 & 39.76 &  30.39 & 23.64& 41.38 & 0.15 \\
% \midrule
% \rowcolor{gray!15}\multicolumn{9}{c}{\textbf{How2Sign-Long}} \\
% Uni-Sign      & 53687 & 53687 & 1.93 & 1.24 & 0.84 & 0.59& 26.32 &0.00  \\
% Uni-Sign(Oracle Reference)      & 53687 & 53687 & 1.93 & 1.24 & 0.84 & 0.59& 26.32 &0.00  \\
% \textbf{Ours}($\beta$=0.2, $k$=1) & 1580  & 3405 & 49.11 & 31.12& 20.97 &14.71 & 38.42 &  0.87\\
% \bottomrule
% \end{tabular}
% \label{tab:complete_long_form}
% \end{table}

\subsection{Quantitative Analysis of Sentence Commitment}
\label{quantitative_commitment}
We further provide a simple quantitative analysis of sentence commitment on the two long-form test sets. Since the model performs online commitment rather than explicit boundary timestamp prediction, we evaluate whether the number of committed sentences roughly matches the number of reference sentences in each long-form video.

\begin{table}[H]
\centering
\caption{Statistics of sentence commitment on the two long-form test sets.}
\label{tab:commitment_statistics}
\begin{tabular}{lcc}
\toprule
Metric & How2Sign-Long & CSL-Story \\
\midrule
Mean Abs. Diff. & 3.48 & 0.83 \\
Max Abs. Diff. & 21 & 3 \\
Std. Dev. & 3.30 & 0.90 \\
\bottomrule
\end{tabular}
\end{table}

% Here, Abs. Diff. denotes the absolute difference between the number of reference sentences and the number of predicted commitments in each long-form video. This analysis evaluates coarse commitment behavior rather than precise segmentation quality. The results show that the commitment mechanism broadly tracks the sentence structure of the input, instead of producing arbitrary or excessively fragmented segmentation. The larger variance on How2Sign-Long is mainly due to its wider variation in sentence lengths and discourse structure.

% We further analyze the impact of silence detection errors on commitment behavior. Under moderate perturbation, this effect remains limited: on CSL-Story, the proportion of predicted silence tokens that lead to erroneous commitment increases only slightly, from 5.22\% to 5.97\% under $\pm 15$-frame perturbation, as is showned in ~\ref{tab:silence_noise_robustness}. More generally, across the two long-form test sets, such tokens account for only 7.67\% of all predicted silence tokens. This limited impact is partly because silence-based commitment is triggered only when the output consists entirely of silence tokens, while Mask-$k$ can suppress premature silence predictions and allow later correction.

In Table~\ref{tab:commitment_statistics}, Abs. Diff. denotes the absolute difference between the number of reference sentences and predicted commitments in each long-form video. This metric provides a coarse measure of commitment behavior rather than precise segmentation quality. The results show that the commitment mechanism broadly follows the sentence structure of the input, instead of producing arbitrary or overly fragmented segments, with larger variance on How2Sign-Long due to its greater diversity in sentence length and discourse structure. We further examine robustness to model silence detection errors: across the two long-form test sets, only 7.67\% of predicted silence tokens lead to erroneous commitments.

\subsection{Extended Ablation Studies}
Table~\ref{tab:complete_long_form_ablation} presents the detailed ablation study results on long-form datasets. Consistent with \cite{Papi_2022}, we prioritize LAAL for latency evaluation in long-form contexts, as the standard AL metric may yield negative values for overly long predictions, rendering it less interpretable.

\subsection{Raw Numerical Results for Trade-off Curves}
\begin{figure}[H]
  \centering
    \includegraphics[width=0.8\linewidth]{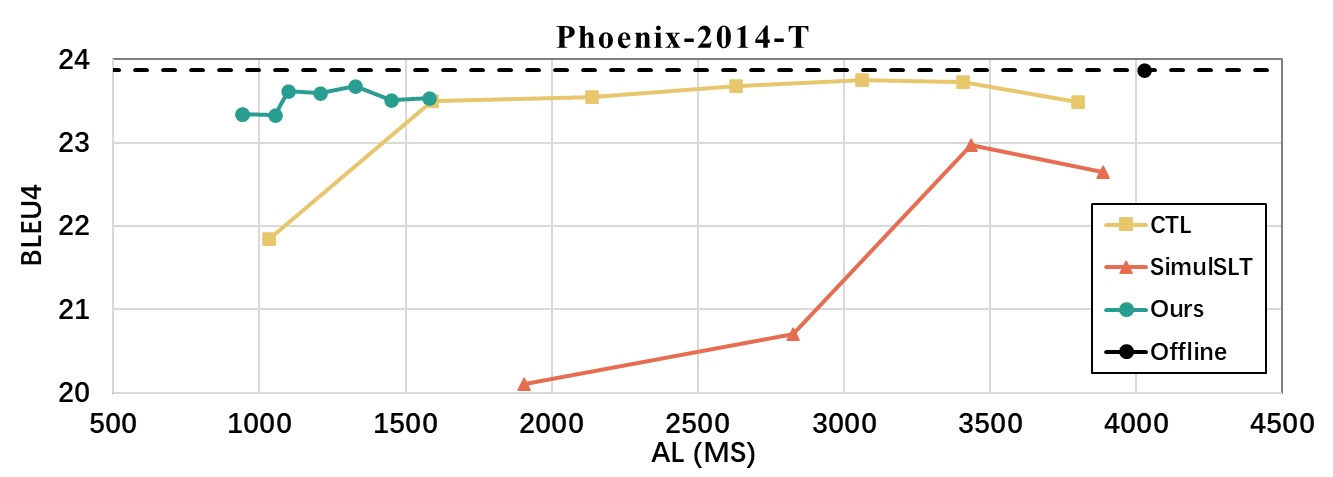 }
  \caption{Sentence-Level Translation quality (BLEU4) vs. latency (Average Lagging) on Phoenix-2014-T. For CTL++, latency is controlled by parameter $m$, with points from left to right corresponding to $m \in \{0, 2, 4, 6, 8, 10, 18\}$. For SimulSLT, latency is adjusted via the wait-$k$ parameter $k_{wait}$, with points from left to right representing $k_{wait} \in \{1, 3, 5, 7\}$. For ReSCUE, latency is adjusted via the Mask-$k$ parameter $k$, with points from left to right representing $k \in \{0, 1, 2, 3, 4, 5, 6\}$. The \textbf{top-left} region represents the optimal performance.
}
  \label{fig:al_bleu4_phx}
\end{figure}

Tables~\ref{abl_csl_daily_bk}, \ref{abl_ph14t_bk}, and ~\ref{abl_h2s_bk} provide the raw numerical data corresponding to the visual comparisons presented in Fig.~\ref{fig:al_bleu4_phx} and the main manuscript. We additionally report the Normalized Erasure (NE) metric to quantify the stability of the streaming output. Furthermore, Table~\ref{abl_longform_bk} details the quantitative results for the long-form datasets under different $k$ settings.

\section{Implementation Details of the SLT Module}

\label{sec:appendix_slt_details}

\subsection{Pose-based Input Representation}
\begin{figure}[h]
    \centering
    \includegraphics[width=0.6\linewidth]{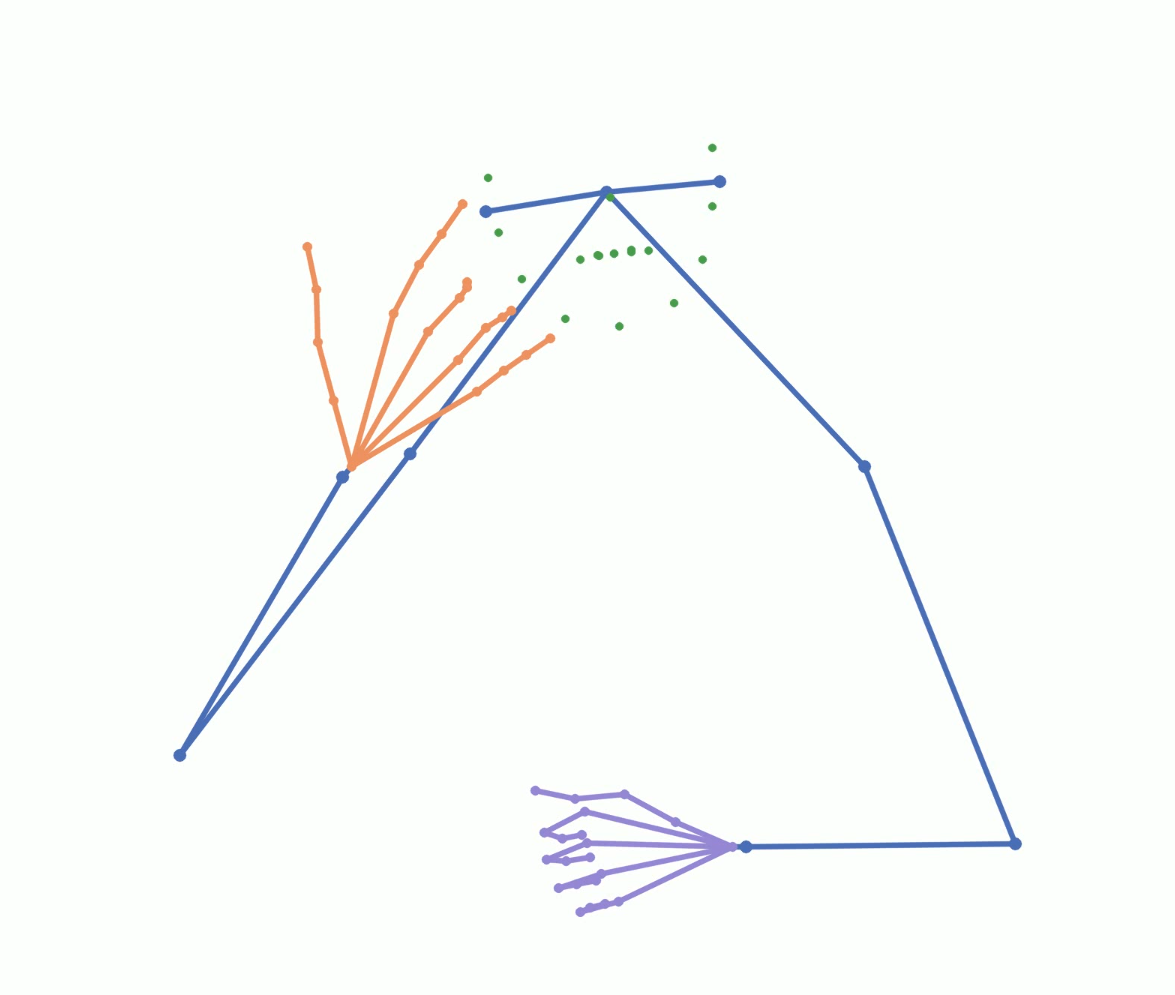} 
    \caption{Illustration of the pose partitioning strategy. The skeleton keypoints are grouped into Face, Hands, and Upper Body patches.}
    \label{fig:pose_partition}
\end{figure}

To achieve real-time inference capability essential for simultaneous translation, we adopt a \textbf{pose-only} visual representation strategy, eschewing computationally intensive RGB inputs. 
Following the formulation in Uni-Sign~\cite{li2025uni}, we extract 2D skeleton keypoints (including face, hands, and upper body) from the raw video stream.
Instead of treating all keypoints as a flat vector, we partition the skeleton into distinct semantic groups to preserve local spatial structural information.
As illustrated in Fig.~\ref{fig:pose_partition}, the keypoints are divided into four semantic patches: \textit{Face}, \textit{Left Hand}, \textit{Right Hand}, and \textit{Upper Body}.
Each partition is independently projected into a feature embedding space before being fed into the Transformer backbone. 
This partitioned modeling allows the network to explicitly focus on fine-grained manual features (hands) and non-manual markers (facial expressions), which are crucial for sign language understanding.

\subsection{Efficiency and Pre-training}
\paragraph{Why Pose-only?}
While RGB-based methods typically yield rich appearance features, they suffer from significant memory footprints and high computational latency, making them ill-suited for the streaming nature of simultaneous translation.
By leveraging the pose-based architecture, our module significantly reduces the input data size and inference time, ensuring low-latency response without compromising linguistic expressiveness.

\paragraph{Pre-training Strategy.}
To bridge the gap between sparse pose inputs and high-quality translation, we initialize our SLT module with the pre-trained weights from the Uni-Sign-Pose variant~\cite{li2025uni}.
Specifically, we utilize language-specific base models to handle the linguistic characteristics of different sign languages:
\begin{itemize}
    \item For the \textbf{Chinese} SLT module, the backbone is pre-trained on the \textbf{CSL-News} dataset, which provides a rich vocabulary for news-domain content.
    \item For the \textbf{English} SLT module, the backbone is initialized with weights pre-trained on \textbf{YouTube-ASL}, ensuring robust generalization across diverse signers.
\end{itemize}
This pre-training paradigm enables the model to learn robust motion primitives and semantic alignments prior to fine-tuning on our simultaneous translation task.

\section{Silence Segments}
\label{sec:vad}

\begin{figure}[t]
  \centering
  
  % (a) Left Figure
  \begin{subfigure}[t]{\linewidth}
    \centering
    \includegraphics[width=0.8\linewidth]{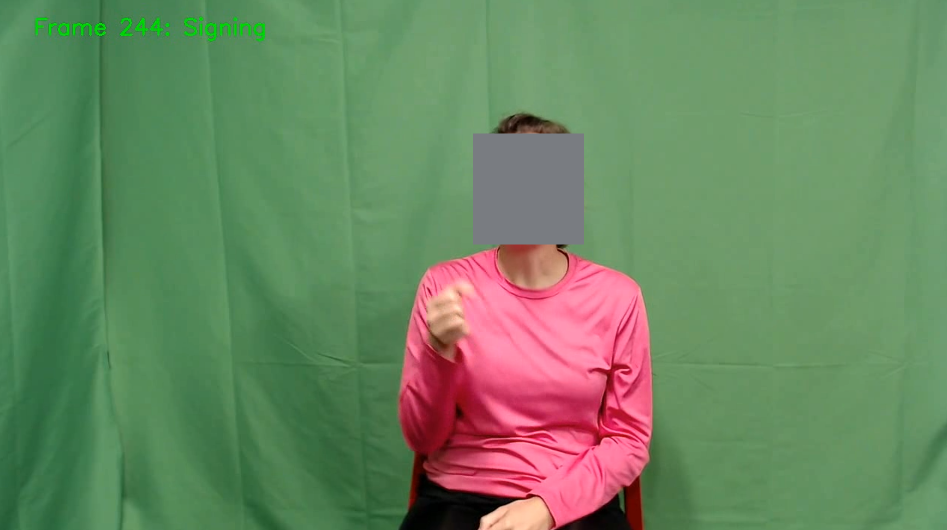}
    \caption{Active frame}
    \label{fig:active}
  \end{subfigure}
  \hfill % Adds horizontal space between the two figures
  % (b) Right Figure
  \begin{subfigure}[t]{\linewidth}
    \centering
    \includegraphics[width=0.8\linewidth]{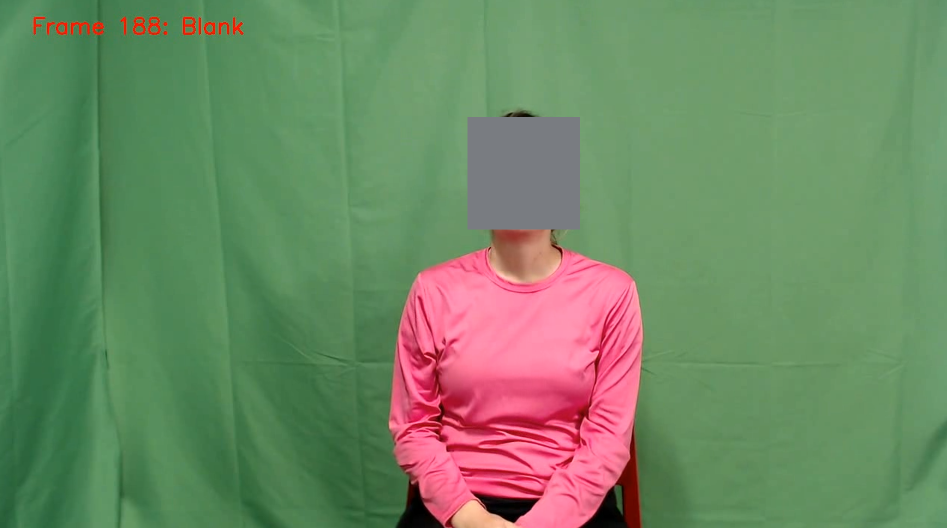}
    \caption{Inactive frame}
    \label{fig:inactive}
  \end{subfigure}

  \caption{Examples of active and inactive frames for silence segmentation.}
  \label{fig:silence_segmentation} % Renamed this label to be unique
\end{figure}

To facilitate the training of the silence segment model, we developed an offline procedure to identify and extract non-signing intervals from the raw video corpus. The determination of a "signing" versus "non-signing" state relies on hand kinematics. Specifically, a frame is classified as \textit{active} (i.e., signing is occurring) if the hand velocity exceeds a speed threshold (i.e., the wrist displacement between consecutive frames exceeds 3\% of the shoulder width) and the vertical hand position surpasses a height threshold (i.e., the hand's vertical position exceeds 65\% of its lowest recorded position). Since the spatial range of motion is analyzed relative to the entire video, this process is performed offline. Fig.~\ref{fig:silence_segmentation} provides examples of active and inactive frames used by this rule.

The detector is used only during training data construction. It is not used at inference time, and does not participate in online translation or sentence commitment.

Based on this detection mechanism, we curate silence segments from two distinct sources:

\begin{enumerate}
    \item \textbf{Global Inactivity:} We extract pure silence segments from parts of the video that exhibit continuous inactivity. A candidate segment is selected if it consists of at least 15 consecutive inactive frames.
    
    \item \textbf{Contextual Padding:} We extract silence segments from the boundaries of annotated sentences. For each sentence, we define a candidate search window of 2 seconds immediately preceding the start and succeeding the end. To ensure label integrity, we first subtract any temporal overlap with adjacent annotated sentences to ensure a ``safe region.'' The extraction logic within this safe region is as follows:
    \begin{itemize}
        \item \textbf{Pre-sentence silence segments:} We examine the safe region before the sentence onset. The silence segment begins at the start of the safe region once at least 3 consecutive inactive frames are detected. The segment extends until the sentence onset unless sustained activity is observed. To suppress transient motion noise, the segment is truncated only when 3 or more consecutive active frames are encountered.
        \item \textbf{Post-sentence silence segments:} We examine the safe region after the sentence end. The silence segment begins at the start of the safe region once at least 3 consecutive inactive frames are detected.  The segment extends to the end of the safe region unless sustained activity is detected. To filter out transient noise, the segment is truncated only if a sequence of 3 or more consecutive active frames is encountered.
    \end{itemize}
\end{enumerate}

\subsection{Threshold Sensitivity}
We evaluate the detector on manually silence-annotated CSL-Story videos by following the same data-construction pattern used in training, i.e., detecting silence around sentence boundaries and measuring the resulting recall. Table~\ref{tab:recall_analysis} shows that the detector is reasonably stable across nearby threshold choices, with the default setting (speed threshold $0.03$, height threshold $0.65$) achieving the highest recall.

\begin{table}[t]
\centering
\caption{Recall of silence detection under different threshold choices on manually annotated CSL-Story videos.}
\vspace{0.5\baselineskip}
\label{tab:recall_analysis}
\begin{tabular}{lccc}
\toprule
Height threshold & \multicolumn{3}{c}{Speed threshold} \\
\cmidrule(lr){2-4}
 & 0.02 & 0.03 & 0.04 \\
\midrule
0.65 & 0.8115 & \textbf{0.8596} & 0.8596 \\
0.85 & 0.7789 & 0.8396 & 0.8396 \\
1.05 & 0.7789 & 0.8396 & 0.8396 \\
\bottomrule
\end{tabular}
\end{table}

\subsection{Full-Video Evaluation}
\begin{figure}[t]
    \centering
    \includegraphics[width=0.8\linewidth]{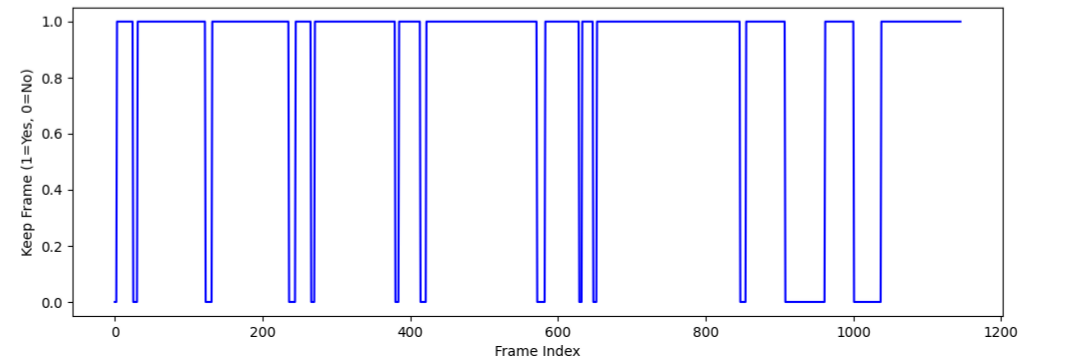}
    \caption{Frame-level detector decisions over time}
    \label{fig:frame_decisions}
\end{figure}

The earlier recall-only analysis was conducted on video regions outside annotated sentence boundaries. In that setting, these regions are treated as silence-only regions, so false positives cannot occur and precision is not informative. To provide a more complete evaluation, we further test the detector on full videos from the manually silence-annotated CSL-Story set, where signing and silence frames must be distinguished jointly. Fig.~\ref{fig:frame_decisions} visualizes the frame-level detector decisions over a full video, illustrating that the detector captures both extended signing spans and intermittent silence intervals.

\begin{table}[t]
\centering
\caption{Frame-level confusion matrix of the silence detector on full CSL-Story videos.}
\vspace{0.5\baselineskip}
\label{tab:silence_confusion_matrix}
\begin{tabular}{lcc}
\toprule
 & Predicted Silence & Predicted Sign \\
\midrule
Actual Silence & 1450 & 545 \\
Actual Sign & 163 & 11422 \\
\bottomrule
\end{tabular}
\end{table}

As shown in the frame-level confusion matrix in Table~\ref{tab:silence_confusion_matrix}, the detector achieves 0.8989 precision, 0.7268 recall, and 0.8038 F1 under full-video evaluation, treating silence as the positive class. These results indicate that the detector is sufficiently reliable for training data construction, while still introducing annotation noise that the downstream model must tolerate.

To assess the impact of imperfect silence annotations, we perturb the sentence--silence boundaries in the CSL-Story training data and retrain the model. The results show that moderate annotation noise causes only limited degradation. Under $\pm 15$-frame perturbation, BLEU-4 drops by 1.51 points, AL increases by 104 ms, and NE increases by 0.08. The effect on commitment behavior is also small: the proportion of predicted silence tokens that lead to erroneous commitments increases only from 5.22\% to 5.97\%. This robustness is partly due to the design of the commitment mechanism, where silence-based commitment is triggered only when the output consists entirely of silence tokens, while Mask-$k$ suppresses premature silence predictions and allows later correction.

% \begin{table}[t]
% \centering
% \caption{Downstream robustness to noisy silence annotations on CSL-Story.}
% \vspace{0.5\baselineskip}
% \label{tab:silence_noise_robustness}
% \begin{tabular}{lccc}
% \toprule
% Boundary noise & BLEU-4 & AL & NE \\
% \midrule
% $\pm 0$ frames & 23.64 & 1588 & 0.15 \\
% $\pm 15$ frames & 22.13 & 1692 & 0.23 \\
% $\pm 25$ frames & 19.83 & 2546 & 0.46 \\
% \bottomrule
% \end{tabular}
% \end{table}

\section{User Study Detail}
\label{sec:user_study}
% ---------------------------------

\subsection{Setup}
In the user study, sign language videos are pre-recorded by signers and processed by three distinct translation systems: ReSCUE, Uni-Sign, and CV-SLT ~\cite{li2025uni,zhao2024conditional}. The translation outputs are integrated into a web-based interface designed for hearing volunteers. The interface displays the source video on the left and the real-time translated captions on the right. Once the participants obtain sufficient information from the translations, they can click the microphone button located beneath the video to provide their responses.

We recruited 4 signing volunteers and 8 hearing volunteers to participate in the user study, comparing it directly against the two different offline methods. 
All participants were adults and provided written informed consent using paper consent forms before participation. For Deaf volunteers, the study instructions and consent information were also communicated in Sign Language to ensure full understanding. 
All participants were compensated at a rate of at least four times the local minimum hourly wage.

Among the signing participants, ages ranged from the 20s to over 40; one was female and three were male. Two were Deaf signers and two were hearing participants with strong sign language proficiency. Among the hearing participants, four self-identified as female and four as male.

\begin{figure*}[t]
  \centering
  \includegraphics[width=0.8\linewidth]{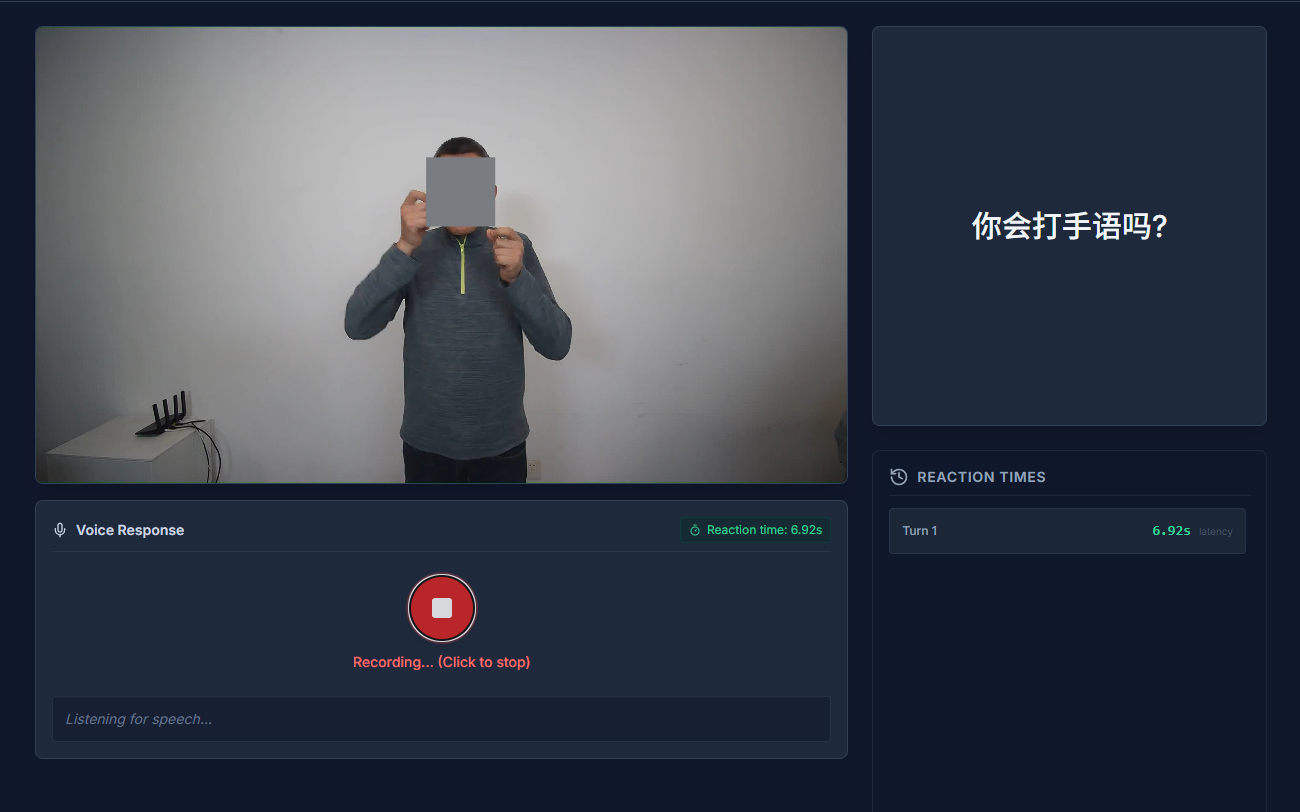}
  \caption{The user interface of Our Simul-SLT system. The simultaneous SLT result is presented on the right of the screen.}
  \label{fig:latency_quality}
\end{figure*}

The study consists of two complementary parts designed to assess both interactive communication and long-form translation capabilities. The user interface of the system is shown in Fig.~\ref{fig:latency_quality}.

\subsection{Sample Composition}
For Part I, the 4 signing participants each recorded 5 videos with 3 sentences in each video, yielding 20 videos and 60 signed sentences. These samples were translated by 3 systems and evaluated by 8 hearing participants, and each response was also rated by the corresponding signer. This results in 960 subjective ratings in total, distributed as 318 for CV-SLT, 318 for Uni-Sign, and 324 for ReSCUE. For Part II, the same participant pool produced 480 trials for objective evaluation.
\subsection{Communication Task Protocol (Part I)}
Deaf signers are instructed to sign sentences on everyday topics.  In each session, a hearing participant viewed these videos and provided a verbal response. Subsequently, a scoring interface appeared, requiring the participant to evaluate the system on a 5-point Likert scale across Translation Quality and Translation Fluency. 
\begin{itemize}
\item \textit{Translation Quality:} Measures whether the intent is correctly conveyed. The Deaf participant rates this based on the relevance and correctness of the hearing partner's responses, while the hearing participant judges whether the translated text can be clearly understood.
\item \textit{Translation Fluency:} Assesses the overall smoothness of interaction, focusing on whether the system enables timely and minimally disruptive communication.
\end{itemize}
\subsection{Long-form Description Task (Part II)}
For each trial, the Deaf signer draws a prompt describing an object or event and produces a continuous sign language description. 
The hearing participant views a multiple-choice question associated with the prompt and selects an answer as soon as the translated content provides sufficient information. 
The web-based interface automatically logs selection accuracy and response time.
\begin{itemize}
\item \textit{Selection Accuracy:} The percentage of correct answers selected by hearing participants, indicating whether the translation supports correct comprehension.
\item \textit{Response Time:} The average elapsed time from the start of signing until the participant selects an answer. Participants are instructed to answer as soon as sufficient information becomes available.
\end{itemize}
This setting evaluates the system's ability to translate continuous sign language input without predefined sentence boundaries.

\subsection{Participant Instructions}

The following instruction text was provided to participants before the study. For Deaf participants, the instructions were also communicated in Sign Language. The wording below is an English translation of the original study instructions.

\paragraph{General instructions.}
You are invited to participate in a study evaluating sign language translation systems. The study aims to understand whether translated captions can support communication between signing and hearing participants. Your participation is voluntary. You may ask questions at any time, skip any trial, or withdraw from the study without penalty. The study does not evaluate your language ability or personal performance. It only evaluates the translation systems.

\paragraph{Instructions for signing participants.}
In Part I, signing participants were given everyday-life topics and were asked to produce three sign-language sentences for each topic. These sentences were designed to cover diverse communicative intents, including descriptions, questions, and other daily conversational expressions. Participants were informed that their signing videos would be translated and that hearing participants would respond based on the translated content. 

In Part II, signing participants were provided with specific prompts, such as describing what they did over the weekend, what they ate recently, how they usually commute, what they like to do in their free time, or what plans they have for an upcoming holiday, and were asked to sign continuously for a period of time, resulting in long-form sign-language videos.

\paragraph{Instructions for hearing participants.}
Hearing participants were informed that they would interact with a web-based system capable of translating sign language. In Part I, hearing participants were asked to communicate with signing participants based on the translations displayed on the webpage. They could press the microphone button at the bottom of the webpage to provide spoken responses to the corresponding topic, which were then converted into text and shown to the signing participants.

In Part II, hearing participants were given a multiple-choice task before the translation started. During real-time translation of the long-form signing video, they were instructed to select an answer as soon as they had obtained sufficient information from the translated content.

\subsection{User Study Results}
Across both studies, ReSCUE achieves higher subjective ratings in quality and fluency (Table~\ref{tab:user_study}(a)), as well as high accuracy and faster response times in the long-form task (Table~\ref{tab:user_study}(b)). These results demonstrate the effectiveness of real-world sign language communication.

\begin{table}[H]
\caption{User study results. O-R denotes Oracle Reference. Scores in panel (a) are averaged over participants on a 5-point Likert scale; higher accuracy and lower response time indicate better performance in panel (b).}
\label{tab:user_study}
\vspace{0.5\baselineskip}
\centering
\small

\begin{minipage}[t]{0.45\linewidth}
\centering
\textbf{(a) Deaf--hearing communication}\\[0.4em]
\setlength{\tabcolsep}{2pt}
\begin{tabular}{lcc}
\toprule
\textbf{Method} & Quality $\uparrow$ & Fluency $\uparrow$ \\
\midrule
CV-SLT (O-R)   & 3.1 & 2.6 \\
Uni-Sign (O-R) & 3.5 & 3.2 \\
\textbf{ReSCUE}  & \textbf{4.3} & \textbf{4.4} \\
\bottomrule
\end{tabular}
\end{minipage}
\hfill
\begin{minipage}[t]{0.50\linewidth}
\centering
\textbf{(b) Long-form SLT}\\[0.4em]
\setlength{\tabcolsep}{2pt}
\begin{tabular}{lcc}
\toprule
\textbf{Method} & \textbf{Acc.} (\%) $\uparrow$ & \textbf{Time (s)} $\downarrow$ \\
\midrule
CV-SLT (O-R)   & 72.4 & 22.3 \\
Uni-Sign (O-R) & \textbf{83.9} & 20.5 \\
\textbf{ReSCUE}  & 78.6 & \textbf{18.7} \\
\bottomrule
\end{tabular}
\end{minipage}

\end{table}
\subsection{Statistical Analysis}
% For Part I, the 4 signing participants each recorded 5 videos with 3 sentences in each video, yielding 20 videos and 60 signed sentences. These samples were translated by 3 systems and evaluated by 8 hearing participants, and each response was also rated by the corresponding signer. This results in 960 subjective ratings in total, distributed as 318 for CV-SLT, 318 for Uni-Sign, and 324 for ReSCUE. For Part II, the same participant pool produced 480 trials for objective evaluation.

In the statistical analysis in Table~\ref{tab:user_study_stats}, the participant is treated as the unit of analysis. Specifically, we first aggregate ratings for each participant and system, and then perform paired $t$-tests across participants. This avoids inflating significance by treating repeated ratings from the same participant as independent observations.

\begin{table}[H]
\centering
\small
\setlength{\tabcolsep}{2pt}
\caption{Paired statistical tests using participants as the unit of analysis.}
\label{tab:user_study_stats}
\begin{tabular}{lccc}
\toprule
Metric & CV-SLT vs. ReSCUE $(t/p)$ & Uni-Sign vs. ReSCUE $(t/p)$ & CV-SLT vs. Uni-Sign $(t/p)$ \\
\midrule
Quality & $-4.011/0.002$ & $-2.230/0.048$ &$-0.960/0.358$ \\
Fluency & $-3.996/0.002$ & $-3.937/0.002$ &$-0.992/0.342$\\
Selection Accuracy & $-4.051/0.005$ & $3.919/0.006$ &$-6.021/0.001$\\
Response Time & $4.312/0.004$ & $2.981/0.020$ &$1.629/0.147$\\
\bottomrule
\end{tabular}
\end{table}

Participant-level paired statistical tests show that ReSCUE outperforms both CV-SLT and Uni-Sign in subjective quality and fluency. Moreover, ReSCUE achieves significantly lower response time than both baselines in the long-form task, demonstrating its effectiveness in improving interaction efficiency while maintaining competitive selection accuracy.

% \subsection{Metrics and Data Collection}
% Subjective scores are averaged across participants and sessions. 
% For the long-form task, accuracy is computed as the percentage of correct selections, and response time is measured from the start of signing to the moment of answer selection.

\subsection{Consent, Compensation, and Review Considerations}

All participants provided written informed consent using paper consent forms before the study.
The consent procedure explained the purpose of the study, the tasks to be performed, the approximate duration, the use of recorded signing videos, and interaction logs for research analysis. 
Participants were informed that they could stop participating at any time without penalty. For Deaf participants, the consent information and task instructions were communicated in Sign Language to ensure accessibility.

Participants were compensated at a rate of at least four times the local minimum hourly wage.

The study involved no deception, medical or psychological intervention, collection of sensitive personal attributes beyond aggregate demographic summaries, or manipulation expected to cause physical or psychological harm.
The tasks consisted of recording or viewing sign language videos, reading translated captions, providing responses, and assigning ratings. 
Therefore, the study posed no foreseeable risk beyond ordinary communication and video-recording privacy considerations. 
To mitigate privacy concerns, participant recordings and logs were used only for research analysis, results are reported in aggregate, and example visualizations are anonymized where applicable. 
Under the applicable institutional policy for minimal-risk user studies of system usability and translation quality, this study did not require IRB review.
\section{Long-Form Dataset}
\label{sec:long-form dataset}

\subsection{How2Sign-Long}
\subsubsection{Construction Strategy.} 
We reconstruct \textbf{How2Sign-Long} from the original How2Sign dataset. At the video level, it contains an average 14 sign language sentences per video. Our inference-aware training framework (Section~\ref{sec:fine_tune}) necessitates an asymmetric strategy for data construction.

\subsubsection{Training.} 
We retain all \textbf{31,086} atomic samples from the original training set to ensure precise sentence-level supervision. Furthermore, we augment the training data to explicitly model non-signing intervals and context transitions. Specifically, we identify \textbf{129,140} silence frames (where $T_{video} > T_{signing}$) and construct \textbf{28,935} multi-sentence pairs from adjacent segments. These mutually exclusive augmentation types provide diverse supervision signals derived from the same source material.

\subsubsection{Testing.} 
Our testing strategy simulates real-world streaming by concatenating adjacent sentences into long-form videos. Statistical analysis of the raw footage shows a mean inter-sentence gap of 0.86s ($\sigma=3.22$s). While most sentences are tightly coupled, we observed that gaps exceeding \textbf{2 seconds} often contain unannotated signing or noise. Based on manual verification, we enforce a 2-second threshold for semantic continuity: adjacent sentences within this window are merged, while larger gaps trigger segmentation. This yields \textbf{297} long-form videos (avg. \textbf{53.69s}), establishing a rigorous evaluation set for stability and latency.

To characterize the difficulty of the evaluation set, we further analyze boundary ambiguity and long silence spans. In the How2Sign-Long test set, 55.9\% of adjacent sentence pairs have a boundary gap shorter than 10 frames, indicating that transitions between neighboring sentences are often ambiguous. In addition, 12 silence segments last longer than 10 seconds, with the longest reaching 36 seconds.

\subsection{CSL-Story}
\begin{figure}[h]
    \centering
    \includegraphics[width=0.8\linewidth]{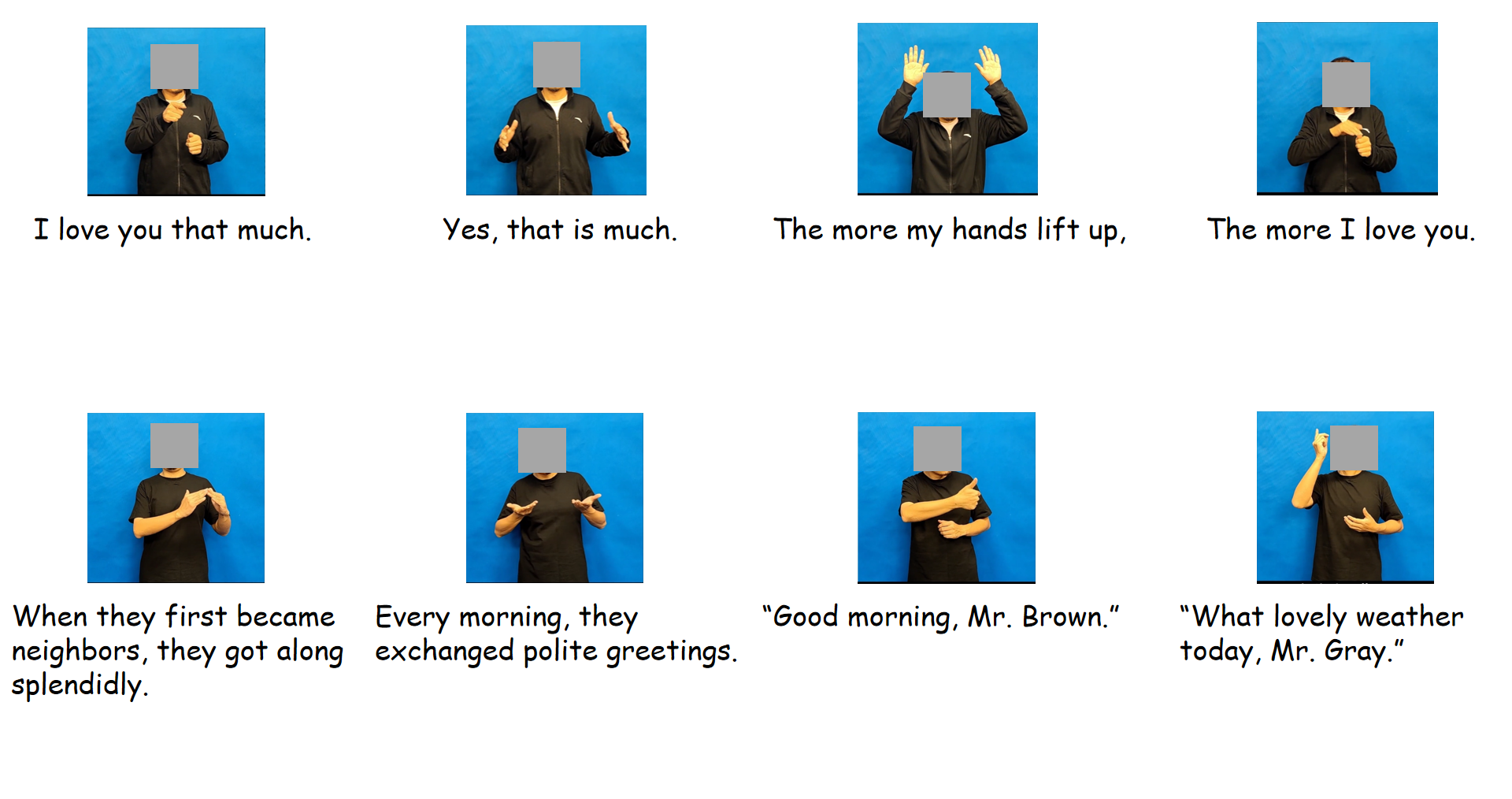}
    \caption{Samples from the CSL-Story Dataset. The dataset features narrative storytelling (e.g., fables) with aligned Chinese subtitles and English translations, introducing vocabulary distinct from daily conversations. The signer's face is masked to protect privacy.}
    \label{fig:csl_story_samples}
\end{figure}

CSL-Story\footnote{The videos are curated from the WeChat public album of \textbf{SuZhou Blind Deaf School}. Link: \url{https://mp.weixin.qq.com/mp/appmsgalbum?__biz=MzA5MDkwODI4OQ==&action=getalbum&album_id=2929466697806381057}} targets narrative domains distinct from daily conversation, introducing an Out-Of-Distribution challenge.
We processed the raw videos using \textbf{Umi-OCR}~\citep{umiocr} to extract subtitles, followed by timestamp-based segmentation. After manual correction of OCR errors to ensure semantic integrity, we obtained \textbf{750} high-quality sentence annotations. At the video level, CSL-Story contains on average 63 sign language sentences per video. Examples from CSL-Story are shown in Fig.~\ref{fig:csl_story_samples}.

\subsubsection{Training.}
Following the How2Sign-Long protocol, the CSL-Story component contributes \textbf{92} atomic narrative samples to the mixed training pool. We further enhance context modeling by incorporating \textbf{583} multi-sentence samples and \textbf{471} silence frames specific to this narrative domain.

\subsubsection{Testing.}
The test set comprises 12 continuous long-form videos. These samples feature high information density, serving as a specialized set for evaluating model stability in uninterrupted narrative streams.

The CSL-Story test set contains challenging temporal structures that make it difficult for models to rely on simple boundary heuristics or overfit to regular sentence patterns. In particular, 34.9\% of adjacent sentence pairs are separated by fewer than 10 frames, creating ambiguous and densely packed boundaries, while 6 silence segments last longer than 10 seconds, with the longest reaching 21 seconds. This combination of near-contiguous sentence transitions and extended non-signing pauses provides a stress test for online segmentation and commitment behavior, making CSL-Story a valuable testing set for evaluating long-form simultaneous SLT in realistic narrative signing.

\subsubsection{Additional Training Analysis}
CSL-Story is relatively small, which raises a potential concern that models evaluated on this dataset may overfit to it.
To reduce this risk, the checkpoint used for CSL-Story experiments in the main paper is jointly trained on CSL-Story and CSL-Daily.
We conduct two additional analyses to examine whether this affects the conclusions.

\paragraph{Evaluation on CSL-Daily.}
We first evaluate the jointly trained checkpoint on CSL-Daily.
As shown in Table~\ref{tab:joint_training_csldaily}, the checkpoint still achieves strong performance on CSL-Daily, with 21.91 BLEU and 49.48 ROUGE.
Compared with the CSL-Daily-only checkpoint reported in Table~\ref{tab:slt_main_csl}, the jointly trained model remains competitive, suggesting that it does not simply overfit to CSL-Story.

\begin{table}[h]
\centering
\small
\caption{Performance on CSL-Daily for the jointly trained checkpoint used in CSL-Story experiments.}
\vspace{0.1in}
\begin{tabular}{lccc}
\toprule
\textbf{Training Set} & AL$\downarrow$ & BLEU$\uparrow$ & ROUGE$\uparrow$ \\
\midrule
CSL-Daily & 2438 & 24.72 & 54.02 \\
CSL-Daily + CSL-Story & 2032.87 & 21.91 & 49.48 \\
\bottomrule
\end{tabular}
\label{tab:joint_training_csldaily}
\end{table}

\paragraph{Training on CSL-Story alone.}
We further examine the effect of training only on CSL-Story, rather than jointly training with CSL-Daily. Table~\ref{tab:story_only_training} shows that story-only training yields higher quality on CSL-Story, suggesting that the CSL-Story is a valid standalone training resource.

\begin{table}[h]
\centering
\small
\caption{Performance on CSL-Story with joint training and CSL-Story-only training.}
\vspace{0.1in}
\begin{tabular}{lccc}
\toprule
\textbf{Training Set} & AL$\downarrow$ & BLEU$\uparrow$ & ROUGE$\uparrow$ \\
\midrule
CSL-Daily + CSL-Story & 1588 & 23.64 & 41.38 \\
CSL-Story & 1830 & 24.91 & 44.89 \\
\bottomrule
\end{tabular}
\label{tab:story_only_training}
\end{table}

\subsection{License}
\textbf{How2Sign.} 
We adhere to the CC BY-NC 4.0 license, utilizing only public RGB videos and keypoint annotations for academic research.

\textbf{CSL-Story.} 
The raw video footage is the intellectual property of \textbf{SuZhou Blind Deaf School}. We do not distribute the processed dataset files directly. 
The raw materials are publicly accessible. The dataset can be reproduced by obtaining the official source and applying the segmentation and processing pipeline detailed in this appendix. We process no personally identifiable information beyond the publicly available visual data.

% \begin{figure}[H]
%   \centering
%   \includegraphics[width=0.8\linewidth]{figs/long-form_demo_v3.pdf}
%   \caption{Demonstration of our simultaneous SLT on long-form input. Grey text denotes tentative predictions being re-translated, while black text represents committed sentences.}
%   \label{fig:simulslt_demo}
% \end{figure}

\section{Dynamic Construction of Inference-Aware Training Instances}
\label{sec:appendix_training_pipeline}

This section details the on-the-fly data construction pipeline used for inference-aware training, which complements the instance types described in Section~\ref{sec:fine_tune}.

We construct a base training pool consisting of $N$ original sentence-level samples and $S$ pure silence samples.
Each sentence-level sample corresponds to a fully annotated signing segment paired with its complete reference translation,
while silence samples consist of non-signing video segments paired with a single \texttt{<SIL>} token.

During training, augmented instances are generated dynamically at each iteration through a probabilistic pipeline controlled by three hyperparameters:
$p_{\text{prefix}}$, $p_{\text{multi}}$, and $p_{\text{silence}}$.
Given a base sample $(\mathbf{X}, \mathbf{Y})$, the following transformations are applied sequentially.
\paragraph{Prefix truncation.}
With probability $p_{\text{prefix}}$, the visual input is truncated to a prefix of the signing segment.
Specifically, a cut point is uniformly sampled within the annotated sentence boundaries,
and both the video and target text are truncated proportionally, following the prefix-to-prefix strategy described in Section~\ref{sec:fine_tune}.
This augmentation exposes the model to incomplete visual evidence and encourages stable partial hypotheses.
\paragraph{Multi-sentence extension.}
For samples that are not truncated by prefix augmentation, we apply multi-sentence extension with probability $p_{\text{multi}}$.
In this case, a prefix of the temporally subsequent sentence is appended to the current sample, including both video and text, and potential silence chunks in between.
This design encourages boundary awareness when the input buffer spans multiple sentences,
while avoiding the requirement to fully translate multiple sentences within a single instance.
\paragraph{Silence augmentation.}
With probability $p_{\text{silence}}$, the temporal boundaries of the video sample are extended to include surrounding non-signing segments,
provided such segments are identified as silence and do not overlap with adjacent annotated sentences.
For prefix-truncated samples, silence extension is restricted to the beginning of the clip to prevent leakage of future signing content.
The corresponding target sequence is augmented with \texttt{<SIL>} tokens to match the number of inserted silence chunks.

These transformations are applied on-the-fly and independently sampled for each training iteration.
As a result, a single base sample may give rise to diverse training instances across epochs,
allowing the model to jointly learn from complete sentences, partial observations, silence intervals,
and ambiguous sentence boundaries under a unified training framework.

\section{Inference Speed and Buffer Analysis}
\label{sec:appendix_inference_speed}

ReSCUE performs re-translation at each inference step, which may raise concerns about computational overhead and potential buffer growth.
In practice, ReSCUE only re-translates the uncommitted active buffer, which is cleared once sentence commitment is triggered.
Therefore, the inference cost is mainly determined by the active buffer length.

\paragraph{Buffer length.}
Since sentence commitment is triggered by terminal punctuation or \texttt{<SIL>}, frequent and reliable commitment is important for preventing buffer overflow.
We analyze the buffer length, measured in frames, at each commitment step on the two long-form datasets.
As shown in Table~\ref{tab:buffer_length}, the buffer length remains well controlled on both datasets.
Even in the worst case, the maximum buffer length is 525 frames.
In practical deployment, a maximum-buffer-length safeguard can be added as a simple engineering fallback.

\begin{table}[h]
\centering
\small
\caption{Buffer length statistics at commitment steps, measured in frames.}
\vspace{0.1in}
\begin{tabular}{lcc}
\toprule
\textbf{Metric} & \textbf{How2Sign-Long} & \textbf{CSL-Story} \\
\midrule
Average & 339.2 & 146.0 \\
Min     & 15    & 30    \\
Max     & 525   & 270   \\
Std     & 180.2 & 72.4  \\
\bottomrule
\end{tabular}
\label{tab:buffer_length}
\end{table}

\paragraph{Runtime.}
We further measure the runtime of each translation step during inference on a single NVIDIA RTX 3090 GPU.
Across the two long-form datasets, the average runtime is 115.42 ms, and the 95th-percentile runtime is 225.00 ms.
Even at the maximum observed buffer length of 525 frames, the runtime is 1248.09 ms.

In our setting, each input chunk contains 15 frames, corresponding to a 500 ms chunk interval for 30 fps videos.
Thus, in most cases, each translation step finishes before the next chunk arrives.
These results show that the computational cost of re-translation is manageable in practice and that ReSCUE can support real-time deployment on a single consumer GPU.
\section{Broader Impacts}
\label{sec:broader_impact}
Simultaneous SLT has the potential to reduce communication barriers between Deaf and hearing communities, especially in interactive settings such as meetings, classrooms, public-service scenarios, and live events.
By operating on continuous unsegmented video streams, ReSCUE moves SLT closer to realistic use cases where pre-segmented sentence clips are unavailable.

At the same time, deployment of automatic SLT systems must be approached with caution.
Translation errors may lead to incomplete or misleading communication, particularly in high-stakes domains such as healthcare, legal services, education assessment, or emergency response. 
Such systems should therefore not be treated as replacements for qualified human interpreters in safety-critical or legally consequential settings without appropriate human oversight.
In addition, SLT systems process videos of people signing, which may contain identifiable biometric and behavioral information.
Responsible deployment should require informed consent,privacy-preserving storage and access control, and clear disclosure that outputs are machine-generated and may be revised or incorrect.
Finally, SLT performance can vary across sign languages, regional variants, signing styles, and signer demographics. 
Future work should include broader community-centered evaluation to avoid unequal performance across signing communities.
\section{Qualitative Evaluation}
\label{sec:failure_cases}
\newcommand{\unstable}[1]{\textcolor{gray}{#1}}
\newcommand{\rawhyp}{\tilde{\mathbf{Y}}_t}
\newcommand{\stablehyp}{\mathbf{Y}_t}
We provide some successful cases and representative failure cases to better understand how the ReSCUE performs in simultaneous long-form SLT. Gray tokens in the ``[ ]'' denote the unstable suffix removed by Mask-$k$.

\subsection{Successful Cases}

\begin{table}[h]
\centering
\small
\setlength{\tabcolsep}{4pt}
\renewcommand{\arraystretch}{1.15}
\begin{tabular}{p{0.18\linewidth} p{0.76\linewidth}}
\toprule
\textbf{Example 1} & Success: Mask-$k$ Prevents Premature Commitment \\
\midrule
\multicolumn{2}{l}{\textbf{Translation Steps}} \\
\midrule
\multicolumn{2}{c}{
\begin{tabular}{c p{0.18\linewidth} p{0.70\linewidth}}
\textbf{Step} & \textbf{Buffer} & \textbf{Hypothesis} \\
\midrule
$68$ 
& $[58:68]$ 
& ...The mulch side to keep weeds down where I walk [\unstable{.}] \\

$72$ 
& $[58:72]$ 
& ...The mulch side to keep weeds down where I walk and now I'm ready to go [\unstable{in}] \\

$75$ 
& $[58:75]$ 
& ...The mulch side to keep weeds down where I walk and now I'm ready to go into the bead, OK. we're going to use [\unstable{a}] \\
\end{tabular}
} \\
\midrule
\textbf{Reference} & ...The mulch was there to keep the weeds down where I walk and now it's all ready to go into the bed. Ok, then we're going to use a... \\
\bottomrule
\end{tabular}

\vspace{0.5mm}
\caption*{\textbf{Analysis.}
In this example, terminal punctuation is predicted before sufficient visual context is observed at step $68$, but Mask-$k$ hides it from the stabilized output and prevents premature commitment.
After more chunks arrive, the model revises the hypothesis and commits the correct sentence. Here $k=1$.}
\end{table}

\begin{table}[h]
\centering
\small
\setlength{\tabcolsep}{4pt}
\renewcommand{\arraystretch}{1.15}
\begin{tabular}{p{0.18\linewidth} p{0.76\linewidth}}
\toprule
\textbf{Example 2} & Success \\
\midrule
\textbf{Final Translation} & Some of you have to push now for a few months but you can't get off your knees. Hi, I'm Carol Ann with Studio Group X, and I'm going to show you how you can improve your push up when you first start working on your push up. Push you in all four. \\
\midrule
\textbf{Reference} & Some of you have been doing push ups now for several months, but you just can't get off your knees. I am CarolAnn with Studio Group X and I'm going to show you how you can improve your push ups. Well, when you first start working on your push ups you go into all fours. \\
\bottomrule
\end{tabular}
\end{table}

\begin{table}[h]
\centering
\small
\setlength{\tabcolsep}{4pt}
\renewcommand{\arraystretch}{1.15}
\begin{tabular}{p{0.18\linewidth} p{0.76\linewidth}}
\toprule
\textbf{Example 3} & Success \\
\midrule
\textbf{Final Translation} & For our second line, we're just going to connect the first line that we just did with about ten degrees line, ok. 
The line goes up, you want to go up about the top of the needle, you don't want to go any further than that if you want your line, you can go up and high on the line of each nail. \\
\midrule
\textbf{Reference} & For our second line, we're going to want to connect the first line that we just did with about a hundred and ten degree line going up. 
You only want to go up to about the top quarter of you nail, you don't want to go all the way to the top.  \\
\bottomrule
\end{tabular}
\end{table}

\newpage
\subsection{Failed Cases}

\begin{table}[h]
\centering
\small
\setlength{\tabcolsep}{4pt}
\renewcommand{\arraystretch}{1.15}
\begin{tabular}{p{0.18\linewidth} p{0.76\linewidth}}
\toprule
\textbf{Example 4} & Error: Premature commitment \\
\midrule
\multicolumn{2}{l}{\textbf{Translation Steps}} \\
\midrule
\multicolumn{2}{c}{
\begin{tabular}{c p{0.18\linewidth} p{0.70\linewidth}}
\textbf{Step} & \textbf{Buffer} & \textbf{Hypothesis} \\
\midrule
$2$ 
& $[1:2]$ 
& It's a kick ok. Hi, this [\unstable{is}] \\

$3$ 
& $[3:3]$ 
& It's a kick ok. OK [\unstable{.}] \\

$4$ 
& $[3:4]$ 
& It's a kick ok. Hi, this is Sean [\unstable{Doo}] \\
\end{tabular}
} \\
\midrule
\textbf{Reference} & The Kick, Ostenado going. So for the Rim clicks we got the two, the and of three... \\
\bottomrule
\end{tabular}

\vspace{0.5mm}
\caption*{\textbf{Analysis.}
The model misinterprets the beginning of a last name starting with ``O'' as ``OK'', which is often followed by terminal punctuation.
This causes an early period prediction and triggers premature commitment at step $2$.
Once committed, the error cannot be revised and further affects subsequent translation.}
\end{table}

\begin{table}[h]
\centering
\small
\setlength{\tabcolsep}{4pt}
\renewcommand{\arraystretch}{1.15}
\begin{tabular}{p{0.18\linewidth} p{0.76\linewidth}}
\toprule
\textbf{Example 5} & Error: Delayed Commitment \\
\midrule
\multicolumn{2}{l}{\textbf{Translation Steps}} \\
\midrule
\multicolumn{2}{c}{
\begin{tabular}{c p{0.18\linewidth} p{0.70\linewidth}}
\textbf{Step} & \textbf{Buffer} & \textbf{Hypothesis} \\
\midrule
$36$ 
& $[15:36]$ 
& ...try to move forward, but if you're set at the basic position, you're always going to want to move [\unstable{.}] \\

$49$ 
& $[15:49]$ 
& ...try to move forward, and if you're going to put your foot on the ground, you always want to move forward, if you [\unstable{are}] \\

$50$ 
& $[15:50]$ 
& ...try to move forward, and if you're going to put your foot on the ground, you always want to move forward. There we go [\unstable{.}] \\
\end{tabular}
} \\
\midrule
\textbf{Reference} & ...trying to be moving forward. If you're in your basic boxing stance and you want to move forward, if you move that back foot first, you're going to cross those legs...  \\
\bottomrule
\end{tabular}

\vspace{0.5mm}
\caption*{\textbf{Analysis.}
From steps 36 to 49, the model does not produce the following sentence, so the terminal punctuation remains in the suffix and is repeatedly hidden by Mask-$k$.
As a result, commitment is delayed and the input buffer continues to grow.
Although the earlier hypothesis contains partially correct information such as ``basic position'', this content is overwritten by later re-translation and disappears from the final committed output.}
\end{table}

\begin{table}[h]
\centering
\small
\setlength{\tabcolsep}{4pt}
\renewcommand{\arraystretch}{1.15}
\begin{tabular}{p{0.18\linewidth} p{0.76\linewidth}}
\toprule
\textbf{Example 6} & Error: Silence Detection Errors \\
\midrule
\multicolumn{2}{l}{\textbf{Translation Steps}} \\
\midrule
\multicolumn{2}{c}{
\begin{tabular}{c p{0.18\linewidth} p{0.70\linewidth}}
\textbf{Step} & \textbf{Buffer} & \textbf{Hypothesis} \\
\midrule
$2$ 
& $[1:2]$ 
&  \texttt{<SIL>} [\unstable{\texttt{<SIL>}}] \\

$3$ 
& $[3:3]$ 
& We're going to do [\unstable{t}] \\

$4$ 
& $[3:4]$ 
& You can see it is a little bit underneath, so [\unstable{I}] \\
\end{tabular}
} \\
\midrule
\textbf{Reference} & When I lock underneath my legs, I'm going to squeeze in. Let me demonstrate you...  \\
\bottomrule
\end{tabular}

\vspace{0.5mm}
\caption*{\textbf{Analysis.}
At step 2, the model incorrectly predicts \texttt{<SIL>} and triggers a commitment, causing meaningful signing frames to be discarded.
Subsequent translation therefore starts after the true sentence beginning, leading to incorrect subjects and sentence openings.
Although later hypotheses may contain some correct words, the missing initial visual context prevents the full sentence from being recovered.}
\end{table}

\newpage
\label{sec:numeric_results}
\FloatBarrier

\begin{table}[H]
\centering
\caption{Comprehensive SLT performance on CSL-Daily. SLRT method is reproduced by \cite{zhou2023gloss}}.
\vspace{0.5\baselineskip}
\resizebox{\linewidth}{!}{
\begin{tabular}{l cccccc cccccc}
\toprule
\multirow{2}{*}{Method}   & \multicolumn{6}{c}{Dev} & \multicolumn{6}{c}{Test} \\
\cmidrule(lr){2-7}\cmidrule(lr){8-13}
& AL(ms)$\downarrow$ & BLEU1$\uparrow$& BLEU2$\uparrow$& BLEU3$\uparrow$ & BLEU4$\uparrow$ & ROUGE$\uparrow$ & AL(ms)$\downarrow$ & BLEU1$\uparrow$& BLEU2$\uparrow$& BLEU3$\uparrow$ & BLEU4$\uparrow$ & ROUGE$\uparrow$ \\
\midrule
\rowcolor{gray!15}\multicolumn{13}{c}{\textbf{offline SLT}}  \\
SLRT \cite{Camgoz_2020_CVPR}      & 4083 & 21.03 & 9.97 &5.96 &4.04 &20.51 &4259 & 20.00 & 9.11 & 4.93 & 3.03 & 19.67 \\
GASLT \cite{yin2023gloss}    &\same & -  & - & - & - & - & \same &19.90 &   9.94 &  5.98& 4.07 & 20.35 \\
GFSLT-VLP \cite{zhou2023gloss} & \same & 39.20  & 25.02  & 16.35& 11.07 & 36.70 &\same & 39.37  &24.93 & 16.26& 11.00 & 36.44 \\
FLa-LLM   \cite{chen2024factorized} & \same & -  &- & -& - & - & \same &37.13 & 25.12 & 18.38& 14.20 & 37.25 \\
Sign2GPT  \cite{wong2024sign2gpt}& \same & - & -  & -&- & - & \same &41.75  & 28.73 & 20.60& 15.40 & 42.36 \\
SignLLM   \cite{gong2024llms} & \same &  42.45 & 26.88 & 17.90 &  12.23 & 39.18 & \same &39.55  & 28.13 & 20.07 & 15.75 & 39.91   \\
SignCL   \cite{ye2024improving} & \same &  - & - & - &  - & - & \same &47.47  &  32.53 & 22.62 & 16.16  &   48.92\\
MSLU \cite{zhou2025scaling}     &  \same & 33.28  & 21.31 & - & 10.27 & 33.13 & \same &33.97 &22.20 & - & 11.42 & 33.80 \\
C$^2$RL  \cite{chen2025c} & \same &  -  & -& -& - & - & \same &49.32  & 36.28 & 27.54 & 21.61 & 48.21  \\
CV-SLT \cite{zhao2024conditional} & \same &  58.05  & 44.73 & 35.14 & 28.24 & 56.36 & \same &58.29  &45.15 &
35.77 & 28.94 & 57.06  \\
Uni-Sign \cite{li2025uni} &   \same     & 53.24 & 40.54 & 31.65 & 25.27 & 54.34 & \same &53.86  & 42.14 & 32.98 & 25.61 & 54.92\\
\midrule
\rowcolor{gray!15}\multicolumn{13}{c}{\textbf{simultaneous SLT}} \\
SimulSLT($k$=7) \cite{simulslt} &  -     & -  &  -& - & - & - & 3370 & 41.52 &28.41 & 19.60 & 13.88 &  43.18 \\
SimulSLT($k$=1)  &   -    & - & - & -& - & - & 1427    &41.19  &27.15 & 18.09& 12.30 &  40.50 \\
CTL++($m$=6) \cite{sun-etal-2024-adaptive} &  -   & -  & -& -& - & - &  2427   &43.72 &29.87 & 20.48 & 14.46 & 41.68 \\
CTL++($m$=0)  &  -   & - &  & & - & - &  1163   & 43.62 &29.65 & 20.15& 14.13 & 41.62 \\
Zuo et al., 2024 \cite{zuo2024towards} &   -   & - &- & - & 21.20 & - & - &  -& -&  - &20.63 & -\\
% ...
\midrule
\textbf{ReSCUE}($\beta$=0.2,$k$=0)  &  680    & 50.83  & 37.41 &28.53  & 22.34 & 50.77 &693    & 51.84  &38.58 & 29.67 & 23.32 & 51.67 \\
\textbf{ReSCUE}($\beta$=0.2,$k$=6)  &  2403    & 53.26 & 39.80 & 30.58  & 24.12 & 53.52 &2438    & 53.56  &40.27 &31.24 & 24.72 & 54.02 \\
\bottomrule
\end{tabular}
}
\label{tab:complete_daily}
\end{table}

\begin{table}[H]
\centering
\caption{Comprehensive SLT performance on Phoenix-2014-T}
\vspace{0.5\baselineskip}
\resizebox{\linewidth}{!}{
\begin{tabular}{l cccccc cccccc}
\toprule
\multirow{2}{*}{Method}   & \multicolumn{6}{c}{Dev} & \multicolumn{6}{c}{Test} \\
\cmidrule(lr){2-7}\cmidrule(lr){8-13}
& AL(ms)$\downarrow$ & BLEU1$\uparrow$  & BLEU2$\uparrow$  & BLEU3$\uparrow$ & BLEU4$\uparrow$ & ROUGE$\uparrow$ & AL(ms)$\downarrow$ & BLEU1$\uparrow$  & BLEU2$\uparrow$  & BLEU3$\uparrow$ & BLEU4$\uparrow$ & ROUGE$\uparrow$ \\
\midrule
\rowcolor{gray!15}\multicolumn{13}{c}{\textbf{offline SLT}}  \\
NSLT  \cite{camgoz2018neural}    & 4299     &  31.87 & 19.11 & 13.16 & 9.94 & 31.80  & 4027 & 32.24 &19.03 &12.83 &9.58 & 31.80 \\
GASLT \cite{yin2023gloss}    &  \same       & -  & -&- & - & - & \same     & 39.07 &26.74 & 21.86&  15.74 & 39.86 \\
GFSLT-VLP \cite{zhou2023gloss} & \same        & 44.08 & 33.56 & 26.74 &  22.12 & 43.72  & \same    & 43.71 & 33.18& 26.11& 21.44 & 42.49 \\
FLa-LLM   \cite{chen2024factorized} & \same        & -  & -&- & - & - &  \same    & 45.27 & 35.33 & 28.03 &  28.03 & 23.09 \\
SignCL   \cite{ye2024improving} & \same        & -  & -&- & - & - & \same     & 49.76 &36.85 & 29.97& 22.74 & 49.04  \\
Sign2GPT  \cite{wong2024sign2gpt} & \same       & -  & -&- & - & - &\same     & 49.54 & 35.96 & 28.83 &22.52 & 48.90 \\
SignLLM  \cite{gong2024llms}   & \same         &  46.88 &  33.56 &  26.74 & 25.25 & 47.23 &\same     &  45.21 & 33.18 & 26.11 & 23.40 & 44.49   \\
MSLU \cite{zhou2025scaling}     & \same   & 44.83 & 33.56  & -&20.62 & 46.65 & \same     & 46.56 & 34.21 & -&22.24 & 46.73 \\
C$^2$RL   \cite{chen2025c} & \same        &  - & - &- &- & - &\same      & 52.81 &40.20 &32.20&26.75 & 50.96  \\
CV-SLT \cite{zhao2024conditional} & \same &  55.09 &  42.60 & 34.63 &29.10 & 54.43 & \same     & 54.88  & 54.88 & 42.68& 29.27 & 54.33  \\
\midrule
\rowcolor{gray!15}\multicolumn{13}{c}{\textbf{simultaneous SLT}} \\
SimulSLT($k$=7) \cite{simulslt} &  -  &   47.76 & 35.33 & 27.85& 22.85 & 49.21  & 3885 & 47.18  & 34.85 & 27.47 & 22.65 & 49.47 \\
SimulSLT($k$=1) &  -  &     - & - &- &- & -  & 1905 &  45.69  & 32.66 &24.94 & 20.10 & 45.62 \\
% SignStreamNet \cite{ahmed2025signstreamnet} &    & 36.56 & 15.37 & 38.86 & 35.70 & 14.26 & 36.56 \\
CTL++($m$=8) \cite{sun-etal-2024-adaptive} & -  &     - & - &- &- & -  & 3061 & 49.54 & 36.66 &28.80 & 23.75 & 48.34\\
CTL++($m$=0)  &    -     & - & -&- & - & - & 1033 & 48.57 & 35.28 & 27.12 & 21.84 & 46.94\\
Zuo et al., 2024 \cite{zuo2024towards} &   -   & - &- &- & 23.75 & - &  - & - & - & - & 23.69 & -\\
% ...
\midrule
\textbf{ReSCUE}($\beta$=0.2,$k$=0)  &    957    & 45.38 & 34.61 &27.67 &22.90 & 46.55 & 941 & 45.55 & 35.18 & 28.13&23.34 & 45.73 \\
\textbf{ReSCUE}($\beta$=0.2,$k$=4)  &    1376    & 45.58 & 34.71 &27.70 &22.84 & 46.54 & 1327 & 45.83 & 35.42 & 28.44 &23.68 & 46.26 \\
\bottomrule
\end{tabular}
}
\label{tab:complete_phx}
\end{table}

\begin{table}[H]
\centering
\caption{Comprehensive SLT performance on How2Sign}
\vspace{0.5\baselineskip}
\resizebox{\linewidth}{!}{
\begin{tabular}{l cccccc }
\toprule
\multirow{2}{*}{Method}    & \multicolumn{6}{c}{Test} \\
\cmidrule(lr){2-7}
& AL(ms)$\downarrow$ & BLEU1$\uparrow$ & BLEU2$\uparrow$ & BLEU3$\uparrow$ & BLEU4$\uparrow$ & ROUGE$\uparrow$ \\
\midrule
\rowcolor{gray!15}\multicolumn{7}{c}{\textbf{offline SLT}}  \\
YouTube-ASL\cite{uthus2023youtube}      & 6659 & 37.82 & 24.13 & 16.92 &12.39 &  -\\

GloFE-VN\cite{lin2023gloss}     & \same &  14.94 & 7.27&  3.93 &  2.24 & 12.61 \\
SSVP-SLT\cite{rust2024towards}     & \same & 43.2  & 28.8 & 20.8 & 15.5 & 38.4 \\
C$^2$RL   \cite{chen2025c}      & \same  &29.07 & 18.56 & 12.92 & 9.37 & 27.02 \\
FLa-LLM\cite{chen2024factorized}    & \same & 29.81 &  18.99 &13.27& 9.66 & 27.81 \\
VAP \cite{jiao2024visual}   & \same & 39.22  & - &- & 12.87 & 27.77 \\
SignMusketeers \cite{gueuwou2025signmusketeers}   & \same &  41.5 & 27.2 & 19.3 & 14.3 & - \\
LiTFiC \cite{jang2025lost}   & \same &  - & - & - & 12.7 & 32.5 \\
Uni-Sign \cite{li2025uni}   & \same &  40.2 &27.1 & 19.7& 14.9 & 36.0 \\
\midrule
\rowcolor{gray!15}\multicolumn{7}{c}{\textbf{simultaneous SLT}} \\
\textbf{ReSCUE}($\beta$=0.2,$k$=0)  &  1260    & 39.18  & 26.67&19.64 & 15.03 & 34.59 \\
\textbf{ReSCUE}($\beta$=0.2,$k$=3) &  1457    & 40.07 &27.23  & 19.97&  15.22 & 34.86 \\
\bottomrule
\end{tabular}
}
\label{tab:complete_how2sign}
\end{table}

\begin{table}[H]
\centering
\caption{Comprehensive SLT performance on long-form dataset}
\vspace{0.5\baselineskip}
\setlength{\tabcolsep}{2pt}
\begin{tabular}{l cccccccc}
\toprule
\textbf{Model} & AL$\downarrow$ & LAAL$\downarrow$ & BLEU1$\uparrow$ &BLEU2$\uparrow$ &BLEU3$\uparrow$ &BLEU4$\uparrow$ &Rouge$\uparrow$ & NE$\downarrow$ \\
\midrule
\rowcolor{gray!15}\multicolumn{9}{c}{\textbf{CSL-Story}} \\
Uni-Sign       & 37722 & 37722 & 4.22 &3.31 &2.67 & 2.10 & 32.43 & 0.00 \\
Uni-Sign (Oracle reference)       & 37722 & 37722 & 48.07 & 38.12 &31.54 & 26.63 & 44.76 & 0.00 \\
\textbf{ReSCUE}($\beta$=0.2, $k$=3)        & 1588 & 2725 & 54.48 & 39.76 &  30.39 & 23.64& 41.38 & 0.15 \\
\midrule
\rowcolor{gray!15}\multicolumn{9}{c}{\textbf{How2Sign-Long}} \\
Uni-Sign      & 53687 & 53687 & 1.93 & 1.24 & 0.84 & 0.59& 26.32 &0.00  \\
Uni-Sign (Oracle Reference)      & 53687 & 53687 & 40.97 & 27.85 & 20.49 & 15.59 & 39.61 &0.00  \\
\textbf{ReSCUE}($\beta$=0.2, $k$=1) & 1580  & 3405 & 49.11 & 31.12& 20.97 &14.71 & 38.42 &  0.87\\
\bottomrule
\end{tabular}
\label{tab:complete_long_form}
\end{table}

\begin{table}[t]
\centering
\caption{Ablation study on long-form dataset. Following \cite{Papi_2022}, we prioritize LAAL for latency evaluation, as the standard AL metric may result in negative values for extended long-form inputs.}
\vspace{0.5\baselineskip}
\setlength{\tabcolsep}{3.3pt}
\begin{tabular}{l cccccccc}
\toprule
\textbf{Model} & AL$\downarrow$ & LAAL$\downarrow$ & BLEU1$\uparrow$ &BLEU2$\uparrow$ &BLEU3$\uparrow$ &BLEU4$\uparrow$ &Rouge$\uparrow$ & NE$\downarrow$ \\
\midrule
\rowcolor{gray!15}\multicolumn{9}{c}{\textbf{CSL-Story}} \\
\textbf{ReSCUE}($\beta$=0.2, $k$=3)        & 1588 & 2725 & 54.48 & 39.76 &  30.39 & 23.64& 41.38 & 0.15 \\
\midrule
w/o Sentence Commit  & 10093 & 10431 & 16.16  &12.61 &10.05 & 7.91& 38.89 & 1.01 \\
w/o Silence Segments & -473 & 1699 & 48.34& 33.99& 25.29& 19.26& 37.43 & 0.32 \\
w/o Prefix-to-Prefix & -2470 & 1674 & 47.73 & 35.57 & 27.65 & 21.89 & 41.21 & 0.31 \\
w/o Multi-sentence   & 13525 & 13663 & 21.12 & 16.11 & 12.74 & 9.85& 38.67  & 0.76 \\
\midrule
\rowcolor{gray!15}\multicolumn{9}{c}{\textbf{How2Sign-Long}} \\
\textbf{ReSCUE}($\beta$=0.2, $k$=1)        &  1580  & 3405 & 49.11 & 31.12& 20.97 &14.71 & 38.42 &  0.87 \\
\midrule
w/o Sentence Commit  & 18225 & 18247 & 2.15 & 1.39 & 0.97& 0.69& 27.12 & 1.95 \\
w/o Silence Segments &  6139 & 6503 &42.56&27.33&18.68 & 13.33 &37.23&  0.97\\
w/o Prefix-to-Prefix & 6155 & 6569 & 43.70 & 27.83& 18.92 &13.45 &38.00  & 1.26 \\
w/o Multi-sentence   &6510& 6858 & 42.10 & 26.95& 18.31& 12.98& 37.05 & 1.10 \\
\bottomrule
\end{tabular}
\label{tab:complete_long_form_ablation}
\end{table}

\begin{table}[H]
\vspace{-1\baselineskip}
    \centering
    \caption{Numeric results on Phoenix-2014-T}
    \vspace{0.5\baselineskip}
    \small % 减小字号以适应页面宽度
    \begin{tabular}{l llllllll}
        \toprule
        {SimulSLT} & & & & & & & & \\
        \midrule
        $k$ & 1 & 3 & 5 & 7 & & & & \\
        AL & 1905 & 2826 & 3435 & 3885 & & & & \\
        BLEU1 & 45.69 & 45.89 & 47.34 & 47.18 & & & & \\
        BLEU2 & 32.66 & 33.03 & 35.20 & 34.85 & & & & \\
        BLEU3 & 24.94 & 25.46 & 27.74 & 27.47 & & & & \\
        BLEU4 & 20.10 & 20.70 & 22.97 & 22.65 & & & & \\
        ROUGE & 45.62 & 46.91 & 48.36 & 49.47 & & & & \\
        \midrule
        {CTL++} & & & & & & & & \\
        \midrule
        $m$ & 0 & 2 & 4 & 6 & 8 & 10 & 14 & 18 \\
        AL & 1033 & 1592 & 2137 & 2629 & 3061 & 3406 & 3800 & 3960 \\
        BLEU1 & 48.57 & 49.74 & 49.54 & 49.91 & 49.54 & 49.47 & 49.35 & 49.40 \\
        BLEU2 & 35.28 & 36.61 & 36.60 & 36.64 & 36.66 & 36.59 & 36.46 & 36.40 \\
        BLEU3 & 27.12 & 28.67 & 28.65 & 28.73 & 28.80 & 28.74 & 28.56 & 28.48 \\
        BLEU4 & 21.84 & 23.50 & 23.55 & 23.68 & 23.75 & 23.73 & 23.49 & 23.41 \\
        ROUGE & 46.94 & 48.24 & 48.23 & 48.26 & 48.34 & 48.23 & 48.16 & 48.06 \\
        \midrule
        \textbf{ReSCUE}($\beta$=0.2) & & & & & & & & \\
        \midrule
        $k$ & 0 & 1 & 2 & 3 & 4 & 5 & 6 &  \\
        AL & 941 & 1055 & 1099 & 1207 & 1327 & 1451 & 1582 & \\
        BLEU1 & 45.55 & 45.76 & 45.76 & 45.56 & 45.83 & 45.73 & 45.73 & \\
        BLEU2 & 35.18 & 35.28 & 35.39 & 35.27 & 35.42 & 35.28 & 35.34 & \\
        BLEU3 & 28.13 & 28.15 & 28.39 & 28.33 & 28.44 & 28.29 & 28.33 & \\
        BLEU4 & 23.34 & 23.33 & 23.62 & 23.59 & 23.68 & 23.51 & 23.53 & \\
        ROUGE & 45.73 & 45.90 & 45.96 & 45.87 & 46.26 & 46.25 & 46.00 & \\
        NE & 0.69 & 0.69 & 0.55 & 0.44 & 0.38 & 0.30 & 0.25 & \\
        \bottomrule
    \end{tabular}
     \label{abl_ph14t_bk}
\end{table}

\begin{table}[H]
    \centering
    \caption{Numeric results on CSL-Daily}
    \vspace{0.5\baselineskip}
    \resizebox{0.57\linewidth}{!}{ 
    \begin{tabular}{l llllllll}
        \toprule
        {SimulSLT} & & & & & & & & \\
        \midrule
        $k$ & 1 & 3 & 5 & 7 & & & & \\
        AL & 1427 & 2054 & 2756 & 3370 & & & & \\
        BLEU1 & 41.19 & 42.17 & 42.09 & 41.52 & & & & \\
        BLEU2 & 27.15 & 28.45 & 28.65 & 28.41 & & & & \\
        BLEU3 & 18.09 & 19.38 & 19.59 & 19.60 & & & & \\
        BLEU4 & 12.30 & 13.44 & 13.55 & 13.88 & & & & \\
        ROUGE & 40.50 & 42.63 & 43.28 & 43.18 & & & & \\
        \midrule
        {CTL++} & & & & & & & & \\
        \midrule
        $m$ & 0 & 2 & 4 & 6 & 8 & 10 & 14 & 18 \\
        AL & 1163 & 1608 & 2022 & 2427 & 2815 & 3155 & 3677 & 3983 \\
        BLEU1 & 43.62 & 43.38 & 43.43 & 43.72 & 43.72 & 43.77 & 43.81 & 43.72 \\
        BLEU2 & 29.65 & 29.36 & 29.54 & 29.87 & 29.81 & 29.79 & 29.78 & 29.74 \\
        BLEU3 & 20.15 & 19.89 & 20.19 & 20.48 & 20.41 & 20.35 & 20.34 & 20.31 \\
        BLEU4 & 14.13 & 13.94 & 14.27 & 14.46 & 14.42 & 14.34 & 14.33 & 14.30 \\
        ROUGE & 41.62 & 41.40 & 41.41 & 41.68 & 41.72 & 41.63 & 41.78 & 41.70 \\
        \midrule
        {ReSCUE($\beta$=0.0)} & & & & & & & & \\
        \midrule
        AL & 1711 & 1805 & 1963 & 2149 & 2350 & 2573 & 2803 & \\
        BLEU1 & 54.38 & 54.38 & 54.38 & 54.38 & 54.38 & 54.38 & 54.38 & \\
        BLEU2 & 41.15 & 41.15 & 41.15 & 41.15 & 41.15 & 41.15 & 41.15 & \\
        BLEU3 & 32.07 & 32.07 & 32.07 & 32.07 & 32.07 & 32.07 & 32.07 & \\
        BLEU4 & 25.49 & 25.49 & 25.49 & 25.49 & 25.49 & 25.49 & 25.49 & \\
        ROUGE & 55.28 & 55.28 & 55.28 & 55.28 & 55.28 & 55.28 & 55.28 & \\
        NE & 1.79 & 1.39 & 1.04 & 0.78 & 0.58 & 0.43 & 0.31 & \\
        \midrule
        {ReSCUE($\beta$=0.1)} & & & & & & & & \\
        \midrule
        $k$ & 0 & 1 & 2 & 3 & 4 & 5 & 6 &  \\
        AL & 913 & 1080 & 1327 & 1601 & 1882 & 2188 & 2489 & \\
        BLEU1 & 53.17 & 53.11 & 53.37 & 53.33 & 53.19 & 53.81 & 53.67 & \\
        BLEU2 & 39.95 & 39.82 & 40.03 & 40.00 & 39.93 & 40.51 & 40.35 & \\
        BLEU3 & 31.01 & 30.80 & 30.93 & 30.99 & 30.98 & 31.50 & 31.31 & \\
        BLEU4 & 24.57 & 24.29 & 24.36 & 24.50 & 24.53 & 24.97 & 24.78 & \\
        ROUGE & 53.48 & 53.27 & 53.56 & 53.46 & 53.46 & 54.17 & 54.17 & \\
        NE & 0.65 & 0.44 & 0.30 & 0.21 & 0.15 & 0.11 & 0.08 & \\
        \midrule
        \textbf{ReSCUE($\beta$=0.2)} & & & & & & & & \\
        \midrule
        $k$ & 0 & 1 & 2 & 3 & 4 & 5 & 6 &  \\
        AL & 693 & 893 & 1177 & 1474 & 1800 & 2116 & 2438 & \\
        BLEU1 & 51.84 & 51.87 & 52.08 & 52.93 & 52.78 & 53.34 & 53.56 & \\
        BLEU2 & 38.58 & 38.58 & 38.77 & 39.57 & 39.51 & 40.04 & 40.27 & \\
        BLEU3 & 29.67 & 29.69 & 29.83 & 30.59 & 30.59 & 31.09 & 31.24 & \\
        BLEU4 & 23.32 & 23.39 & 23.41 & 24.15 & 24.20 & 24.60 & 24.72 & \\
        ROUGE & 51.67 & 51.81 & 52.20 & 53.02 & 53.10 & 53.82 & 54.02 & \\
        NE & 0.39 & 0.24 & 0.16 & 0.12 & 0.09 & 0.07 & 0.05 & \\
        \midrule
        {ReSCUE($\beta$=0.3)} & & & & & & & & \\
        \midrule
        $k$ & 0 & 1 & 2 & 3 & 4 & 5 & 6 &  \\
        AL & 597 & 806 & 1100 & 1417 & 1753 & 2085 & 2421 & \\
        BLEU1 & 49.72 & 51.08 & 51.50 & 52.22 & 52.68 & 53.43 & 53.26 & \\
        BLEU2 & 36.54 & 37.84 & 38.35 & 38.96 & 39.40 & 40.04 & 40.05 & \\
        BLEU3 & 27.74 & 29.05 & 29.53 & 30.06 & 30.46 & 31.04 & 31.07 & \\
        BLEU4 & 21.58 & 22.80 & 23.25 & 23.67 & 24.07 & 24.59 & 24.57 & \\
        ROUGE & 50.00 & 51.02 & 51.65 & 52.38 & 52.92 & 53.84 & 53.76 & \\
        NE & 0.25 & 0.15 & 0.10 & 0.07 & 0.06 & 0.05 & 0.04 & \\
        \midrule
        {ReSCUE($\beta$=0.4)} & & & & & & & & \\
        \midrule
        $k$ & 0 & 1 & 2 & 3 & 4 & 5 & 6 &  \\
        AL & 488 & 716 & 1042 & 1368 & 1728 & 2069 & 2419 & \\
        BLEU1 & 48.03 & 49.58 & 50.39 & 51.64 & 52.16 & 53.17 & 53.20 & \\
        BLEU2 & 34.48 & 36.18 & 37.15 & 38.48 & 38.92 & 39.81 & 39.97 & \\
        BLEU3 & 25.56 & 27.38 & 28.45 & 29.65 & 30.06 & 30.80 & 31.03 & \\
        BLEU4 & 19.48 & 21.23 & 22.25 & 23.31 & 23.72 & 24.33 & 24.56 & \\
        ROUGE & 47.96 & 49.18 & 50.51 & 51.67 & 52.25 & 53.57 & 53.73 & \\
        NE & 0.14 & 0.07 & 0.05 & 0.04 & 0.04 & 0.04 & 0.04 & \\
        \bottomrule
    \end{tabular}
    }
    \label{abl_csl_daily_bk}
\end{table}

\textbf{
\begin{table}[H]
    \centering
    \caption{Numeric results on How2Sign}
    \vspace{0.5\baselineskip}
    \small % 减小字号以适应页面宽度
    \begin{tabular}{l lllll}
        \toprule
        \textbf{ReSCUE($\beta$=0.2)} & & & & &  \\
        \midrule
        $k$ & 0 & 1 & 2 & 3  \\
        AL & 1260 & 1239 & 1339 & 1457 \\
        BLEU1 & 39.18 & 39.69 & 39.95 & 40.07 \\
        BLEU2 & 26.67 & 27.00 & 27.17 & 27.23 \\
        BLEU3 & 19.64 & 19.86 & 19.96 & 19.97 \\
        BLEU4 & 15.03 & 15.18 & 15.24 & 15.22 \\
        ROUGE & 34.59 & 34.82 & 34.90 & 34.86 \\
        NE & 0.60 & 0.63 & 0.50 & 0.44 \\
        \bottomrule
    \end{tabular}
     \label{abl_h2s_bk}
\end{table}}
\begin{table}[H]
    \centering
    \caption{Numeric results on long-form dataset}
    \vspace{0.5\baselineskip}
    \small % 减小字号以适应页面宽度
    \begin{tabular}{l lllll}
        \toprule
        $k$ & 0 & 1 & 2 & 3  \\
        \midrule
        \rowcolor{gray!15}\multicolumn{6}{c}{\textbf{CSL-Story}} \\
        \midrule
        \textbf{ReSCUE($\beta$=0.2)} & & & & &  \\
        \midrule
        LAAL & 1059 & 1530 & 1869 & 2725 \\ 
        AL & -2041 & -201 & 565 & 1588 \\
        BLEU1 & 48.85 & 51.07 & 51.70 & 54.48 \\
        BLEU2 & 35.62 & 37.93 & 37.69 & 39.76 \\
        BLEU3 & 27.51 & 29.13 & 28.51 & 30.39 \\
        BLEU4 & 21.41 & 22.91 & 22.13 & 23.64 \\
        ROUGE & 36.76 & 39.65 & 41.87 & 41.38 \\
        NE & 0.66 & 0.29 & 0.20 & 0.15 \\
        \midrule
        \rowcolor{gray!15}\multicolumn{6}{c}{\textbf{How2Sign-Long}} \\
        \midrule
        \textbf{ReSCUE($\beta$=0.2)} & & & & &  \\
        \midrule
        LAAL & 824 & 3405 & 3944 & 4348   \\
        AL & -11675 & 1580  & 2651 &  3387   \\
        BLEU1  & 26.24 & 49.11 & 48.31 &   47.47 \\
        BLEU2  & 15.75 & 31.12 & 30.69 &  29.75  \\
        BLEU3 & 9.97 & 20.97 & 20.54 &  19.76  \\
        BLEU4  & 6.51 & 14.71 & 14.41 &  13.74  \\
        ROUGE  & 35.15 & 38.42 & 38.35 &  38.06  \\
        NE  & 0.12 & 0.87 & 0.76 &  0.72  \\
        \bottomrule
    \end{tabular}
     \label{abl_longform_bk}
\end{table}

% \section{Technical appendices and supplementary material}
% Technical appendices with additional results, figures, graphs, and proofs may be submitted with the paper submission before the full submission deadline (see above). You can upload a ZIP file for videos or code, but do not upload a separate PDF file for the appendix. There is no page limit for the technical appendices. 

% Note: Think of the appendix as ``optional reading'' for reviewers. The paper must be able to stand alone without the appendix; for example, adding critical experiments that support the main claims to an appendix is inappropriate. 

%%%%%%%%%%%%%%%%%%%%%%%%%%%%%%%%%%%%%%%%%%%%%%%%%%%%%%%%%%%%

\clearpage
\section*{NeurIPS Paper Checklist}

\begin{enumerate}

\item {\bf Claims}
    \item[] Question: Do the main claims made in the abstract and introduction accurately reflect the paper's contributions and scope?
    \item[] Answer: \answerYes{} % Replace by \answerYes{}, \answerNo{}, or \answerNA{}.
    \item[] Justification: Main claims made in the abstract and introduction accurately reflect the paper's contribution and scope.

\item {\bf Limitations}
    \item[] Question: Does the paper discuss the limitations of the work performed by the authors?
    \item[] Answer: \answerYes{} % Replace by \answerYes{}, \answerNo{}, or \answerNA{}.
    \item[] Justification: The paper discusses the limitations of the work performed by us, in Section 6, Appendix I.
    
\item {\bf Theory assumptions and proofs}
    \item[] Question: For each theoretical result, does the paper provide the full set of assumptions and a complete (and correct) proof?
    \item[] Answer: \answerNA{} % Replace by \answerYes{}, \answerNo{}, or \answerNA{}.
    \item[] Justification: The paper does not include formal proofs. Instead, we provide equations that specify the sentence commitment mechanism and the overall ReSCUE architecture. We state the necessary assumptions and define the relevant notation to make the implementation procedure clear and reproducible.

    \item {\bf Experimental result reproducibility}
    \item[] Question: Does the paper fully disclose all the information needed to reproduce the main experimental results of the paper to the extent that it affects the main claims and/or conclusions of the paper (regardless of whether the code and data are provided or not)?
    \item[] Answer: \answerYes{} % Replace by \answerYes{}, \answerNo{}, or \answerNA{}.
    \item[] Justification: The paper describes the model architecture, inference procedure, training strategy, datasets, evaluation metrics, hyperparameters, and ablation settings in Sections 3--5 and Appendices B, C, D, F, G, and H. These details provide a reasonable path for reproducing the main experimental results.

\item {\bf Open access to data and code}
    \item[] Question: Does the paper provide open access to the data and code, with sufficient instructions to faithfully reproduce the main experimental results, as described in supplemental material?
    \item[] Answer: \answerYes{} % Replace by \answerYes{}, \answerNo{}, or \answerNA{}.
    \item[] Justification: We plan to release the training and inference pipeline, trained models, and documentation upon acceptance, while respecting the licenses and terms of use of the underlying datasets.

\item {\bf Experimental setting/details}
    \item[] Question: Does the paper specify all the training and test details (e.g., data splits, hyperparameters, how they were chosen, type of optimizer) necessary to understand the results?
    \item[] Answer: \answerYes{} % Replace by \answerYes{}, \answerNo{}, or \answerNA{}.
    \item[] Justification: Section 5.1 specifies the datasets, evaluation metrics, optimizer, learning rate, weight decay, GPU setup, batch size, chunk size, biased search weight, and training strategy probabilities. Additional dataset construction and training details are provided in Appendices F and G.

\item {\bf Experiment statistical significance}
    \item[] Question: Does the paper report error bars suitably and correctly defined or other appropriate information about the statistical significance of the experiments?
    \item[] Answer: \answerNo{} % Replace by \answerYes{}, \answerNo{}, or \answerNA{}.
    \item[] Justification: The main automatic SLT results are reported without error bars. For the user study, Appendix E.6 reports participant-level paired statistical tests, but we do not provide statistical significance analysis for all automatic evaluation results.

\item {\bf Experiments compute resources}
    \item[] Question: For each experiment, does the paper provide sufficient information on the computer resources (type of compute workers, memory, time of execution) needed to reproduce the experiments?
    \item[] Answer: \answerYes{} % Replace by \answerYes{}, \answerNo{}, or \answerNA{}.
    \item[] Justification: The paper reports the computer resources used for training and inference, batch size, and inference runtime analysis in Section 5.1 and Appendix H.
    
\item {\bf Code of ethics}
    \item[] Question: Does the research conducted in the paper conform, in every respect, with the NeurIPS Code of Ethics \url{https://neurips.cc/public/EthicsGuidelines}?
    \item[] Answer: \answerYes{} % Replace by \answerYes{}, \answerNo{}, or \answerNA{}.
    \item[] Justification: We reviewed the NeurIPS Code of Ethics and designed the study to respect participant consent, privacy, and appropriate use of public and licensed assets. The user study was conducted with informed consent and compensation, and the dataset release plan avoids distributing private processed video files.

\item {\bf Broader impacts}
    \item[] Question: Does the paper discuss both potential positive societal impacts and negative societal impacts of the work performed?
    \item[] Answer: \answerYes{} % Replace by \answerYes{}, \answerNo{}, or \answerNA{}.
    \item[] Justification: The paper discusses positive societal impacts for Deaf-hearing communication and inclusion, and discusses negative societal impacts in the Appendix I.
    
\item {\bf Safeguards}
    \item[] Question: Does the paper describe safeguards that have been put in place for responsible release of data or models that have a high risk for misuse (e.g., pre-trained language models, image generators, or scraped datasets)?
    \item[] Answer: \answerYes{} % Replace by \answerYes{}, \answerNo{}, or \answerNA{}.
    \item[] Justification: The paper describes privacy-related safeguards for CSL-Story: raw materials remain under the original public source, and sample visualizations mask the signer's face. The released implementation will focus on reproducibility while respecting source licenses and privacy constraints.

\item {\bf Licenses for existing assets}
    \item[] Question: Are the creators or original owners of assets (e.g., code, data, models), used in the paper, properly credited and are the license and terms of use explicitly mentioned and properly respected?
    \item[] Answer: \answerYes{} % Replace by \answerYes{}, \answerNo{}, or \answerNA{}.
    \item[] Justification: The paper cites the existing datasets, models, and tools used in the work, and Appendix F.3 discusses the licenses and ownership of How2Sign and CSL-Story.

\item {\bf New assets}
    \item[] Question: Are new assets introduced in the paper well documented and is the documentation provided alongside the assets?
    \item[] Answer: \answerYes{} % Replace by \answerYes{}, \answerNo{}, or \answerNA{}.
    \item[] Justification: The paper introduces How2Sign-Long and CSL-Story and documents their construction, training/testing splits, statistics, samples, and license/privacy considerations in Appendix F. Appendix A describes the planned release of code, trained models, and documentation.

\item {\bf Crowdsourcing and research with human subjects}
    \item[] Question: For crowdsourcing experiments and research with human subjects, does the paper include the full text of instructions given to participants and screenshots, if applicable, as well as details about compensation (if any)? 
    \item[] Answer: \answerYes{} % Replace by \answerYes{}, \answerNo{}, or \answerNA{}.
    \item[] Justification: Appendix E provides the user study participants, user interface, task protocols, full participant instructions, written informed consent procedure, and compensation.

\item {\bf Institutional review board (IRB) approvals or equivalent for research with human subjects}
    \item[] Question: Does the paper describe potential risks incurred by study participants, whether such risks were disclosed to the subjects, and whether Institutional Review Board (IRB) approvals (or an equivalent approval/review based on the requirements of your country or institution) were obtained?
    \item[] Answer: \answerYes{} % Replace by \answerYes{}, \answerNo{}, or \answerNA{}.
    \item[] Justification: Appendix E describe the consent procedure, privacy considerations, risk assessment, and review considerations. The study involved minimal-risk usability and translation-quality evaluation, and under the applicable institutional policy did not require IRB review.

\item {\bf Declaration of LLM usage}
    \item[] Question: Does the paper describe the usage of LLMs if it is an important, original, or non-standard component of the core methods in this research? Note that if the LLM is used only for writing, editing, or formatting purposes and does \emph{not} impact the core methodology, scientific rigor, or originality of the research, declaration is not required.
    %this research? 
    \item[] Answer: \answerYes{} % Replace by \answerYes{}, \answerNo{}, or \answerNA{}.
    \item[] Justification: The paper describes the use of a multimodal language model as part of the SLT module in Section 3.1 and further details the pose-based input representation and pretrained backbone in Appendix C.

\end{enumerate}

\end{document}